\documentclass[11pt]{article}

\usepackage[final]{acl}

\usepackage{times}
\usepackage{latexsym}

\usepackage[T1]{fontenc}

\usepackage[utf8]{inputenc}

\usepackage{microtype}

\usepackage{inconsolata}

\usepackage{graphicx}

\usepackage{tabularx} 
\usepackage{array}
\usepackage{multirow}
\usepackage{booktabs}
\usepackage{graphicx} 
\usepackage{url} 
\usepackage{amsthm,amsmath,amssymb} 
\usepackage{mathrsfs} 
\usepackage{bm} 
\usepackage{pifont} 
\usepackage{float}
\usepackage{lipsum}
\usepackage{soul}
\usepackage{caption}
\usepackage{subcaption}
\usepackage{enumitem}
\usepackage{pdfpages}
\usepackage{dsfont}
\usepackage{color}
\usepackage{colortbl}
\usepackage{xcolor}
\usepackage{tcolorbox}
\tcbuselibrary{skins,breakable}
\usepackage{tikz}
\definecolor{deepred}{RGB}{180,0,0} 
\definecolor{customred}{RGB}{170, 20, 20} 
\definecolor{customgreen}{RGB}{20, 170, 20} 
\usepackage{xspace}
\usepackage[title]{appendix}

\title{HSS-Synth: Humanities and Social Sciences Data Synthesis for LLMs}

\author{
\textbf{Ru Peng$^{1}$\thanks{This work was done during internships at Qwen Team, Alibaba Group and Inclusion AI, Ant Group.}, \quad
Tianyu Zhao$^{2}$, \quad
Xijun Gu$^{1}$, \quad
Zhiting Fan$^{1}$,} \\[2pt]
\textbf{Haokai Xu$^{1}$, \quad
Jinyang Zhang$^{3,4}$, \quad
Yawen Zeng$^{1}$, \quad
Yihong Zhuang$^{2}$,} \\[2pt]
\textbf{Kexin Yang$^{4}$, \quad
Junyang Lin$^{4}$, \quad
Dayiheng Liu$^{4}$\thanks{Corresponding authors.}, \quad
Junbo Zhao$^{1}$\footnotemark[2]} \\[6pt]
$^1$Zhejiang University \qquad
$^2$Inclusion AI, Ant Group \\[1pt]
$^3$Peking University \qquad
$^4$Qwen Team, Alibaba Group \\[4pt]
\texttt{\{rupeng, j.zhao\}@zju.edu.cn} \qquad
\texttt{liudayiheng.ldyh@alibaba-inc.com}
}

\begin{document}
\maketitle

\begin{abstract}
High-quality, diverse data are vital for large language models (LLMs) but remain scarce and costly. Data synthesis is a viable alternative and succeeds on closed tasks, yet the humanities and social sciences (HSS) are overlooked, and their open-ended nature makes synthesis challenging.
Moving beyond prior capability-centric, fragmented attempts, we adopt a subject-centric paradigm, define the first HSS domain system covering 14 mainstream fields, and introduce HSS-Synth—the first data synthesis pipeline for HSS.
HSS-Synth comprises: (1) constructing seed document from web corpora via multi-step filtering and text refinement evaluated by a judge; (2) specifying “requirements + persona” to backtranslate seed document into diverse yet faithful instructions with strict Q\&A alignment check;  and (3) breaking LLM response limits via teacher-forced Answering that fed seed documents during response to anchor semantics, reduce hallucinations, and preserve tone and integrity.
HSS-Synth yields 237k high-quality, diverse instruction-tuning samples that outperform 14 leading baselines on 16 benchmarks. The fine-tuned Qwen3-8B-Base set new SOTA and approached official Qwen3-8B, improving both human preference and knowledge capability without performance seesaws. 
Extensive experiments demonstrate the HSS-Synth's robustness and transferability.
Our code is publicly available at \url{https://github.com/pengr/HSS-Synth}.
\end{abstract}

\section{Introduction}
The capabilities of large language models (LLMs) hinge on high-quality, diverse data~\cite{peng2025dataman}, yet such data remain scarce and costly~\cite{muennighoff2023scaling, liu2024best}. Consequently, data synthesis has emerged as a viable counterpart~\cite{huang2022large,abdin2024phi}, achieving proven success on closed tasks—mathematics, code, tool use, instruction following, and tabular data~\cite{yu2024metamath, austin2021program, cai2023large, wang2023selfinstruct, zhao2025tabula}. However, the humanities and social sciences (HSS)—the cornerstone of human society and knowledge—have been largely overlooked. HSS, as open-ended tasks lack verifiable answers and demand nuanced judgment, poses significant challenges for data synthesis. Existing synthesis efforts are fragmented and confined to narrow scenarios, such as creative writing, role-playing, dialogue, and long-context~\cite{wang2024weaver, ge2024scaling, qian2025bottom, bai2024longwriter}. Departing from above capability-centric, fragmented efforts and driven by LLM deployment needs~\cite{naveed2023comprehensive}, we adopt a subject-centric paradigm: using the official taxonomy\footnote{\url{https://www.topuniversities.com/subject-rankings}}, we define \textbf{the first HSS domain system with 14 mainstream fields}\footnote{
In this paper, HSS refers to disciplines studying human behavior, cultural expression, social structures, and institutional mechanisms, including Philosophy, Economics, Law, Politics, Sociology, Healthcare, Geography, Education, Sports, Literature, History, Management, Arts, and Psychology.} and \textbf{conduct the first data synthesis work} across HSS domains. A synthetic HSS sample case see Appendix Table~\ref{tab:case_study}.

\begin{figure*}[!t]
\centering
\includegraphics[width=0.96\linewidth]{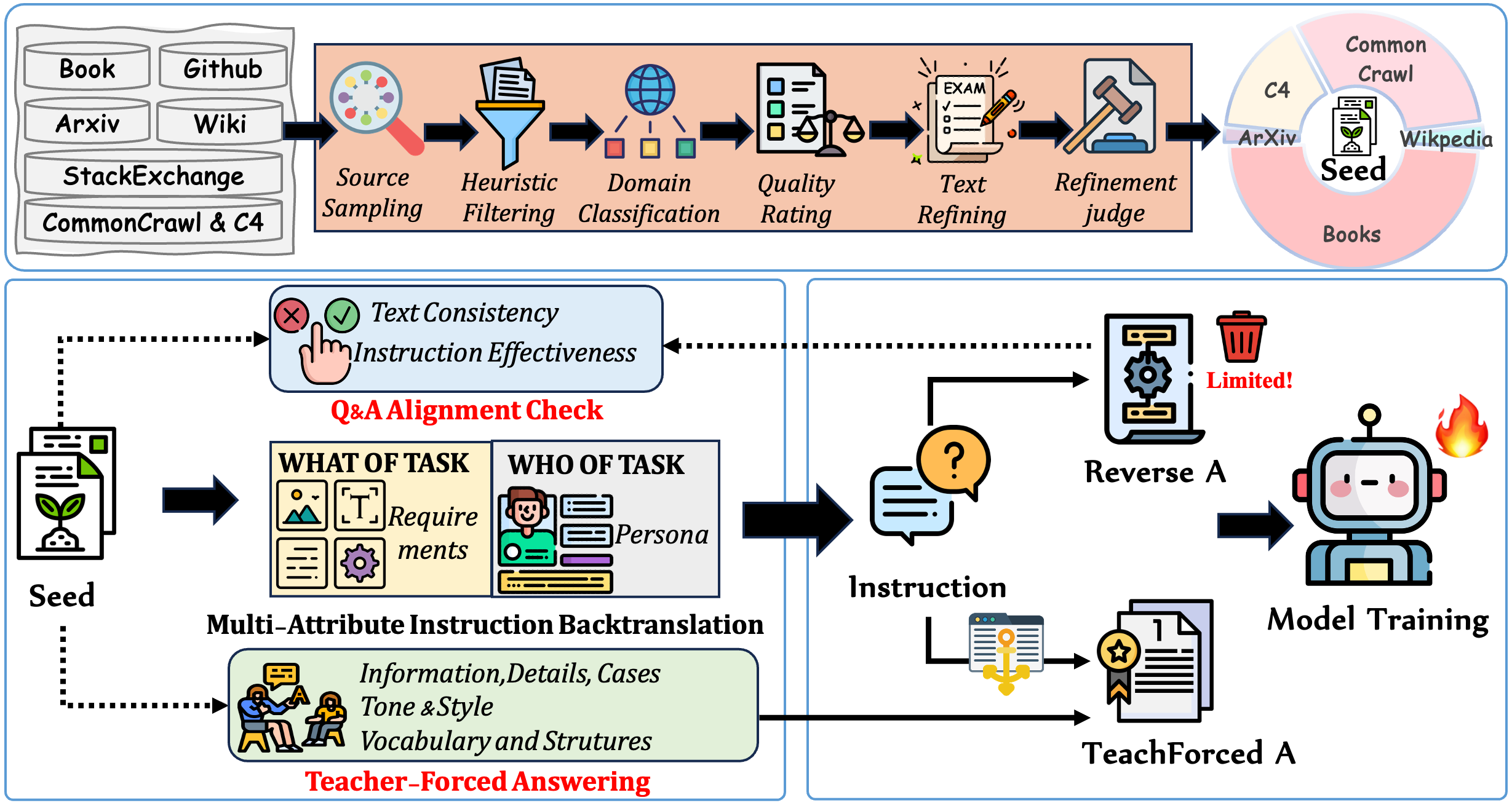}
\caption{
Overview of the \textbf{HSS-Synth} pipeline: (1) Seed document construction: sample, filter, classify, rate, and refine web texts under judge into clean HSS seeds; (2) Question-Answer Pair Refinement—backtranslate instructions from seed with multiple attributes, specify the “what” and “who,” and enforce Q\&A alignment checks to yield diverse, faithful Q\&A pairs; (3) Teacher-forced Answering—anchor answers on seeds to better response beyond the LLM’s inherent limits. Finally, we train models on the resulting instruction-tuning pairs.
}
\vspace{-12pt}
\label{fig:method}
\end{figure*}

Along this line, we propose \textbf{HSS-Synth}, the first data synthesis pipeline for HSS. Figure~\ref{fig:method} illustrates its three stages:
1) \textbf{Seed document construction}: Given abundant scale and diversity of web corpora, we construct seed documents from the web. To address noise and the scarcity of ideal HSS texts, we integrate source sampling, heuristic filtering, domain classification, and quality rating based on 12 expert-crafted rubrics to select HSS documents that satisfy quality standards. To handle residual noise, redundancy, and weak expression beyond filtering, we apply LLM-based text refinement and verify success with a refinement judge, yielding clean seed documents.
2) \textbf{Question-answer pair refinement}: Unlike simple instruction backtranslation~\cite{koksal2023longform,li2024selfalignment}, we adopt multi-attribute instruction backtranslation that specifies both the `what to do'' (requirements) and the` who to do'' (persona), generating diverse yet faithful instructions from seed documents. Followed by a Q\&A alignment check to strictly enforce consistency between instructions and seed documents, further improving instruction fidelity.
3) \textbf{Teacher-forced answering}: Even after Q\&A pair refinement, LLM-generated answers may exhibit factual gaps, detail lacks, and style drift~\cite{cao2025condor,jiang2025instruction}. We therefore propose Teacher-Forced Answering (TeachForceA): during answering, the seed document is fed alongside the instruction as semantic anchors, providing required information to reduce hallucination, preserving tone and style to enhance human-touch, and leveraging structural and lexical cues to improve completeness and readability. This breaks the ceiling of LLM-generated answers.
Using HSS-Synth, we synthesize an instruction-tuning dataset that meets: (i) \emph{sufficient scale} (237k samples); (ii) \emph{high quality} (strong Q\&A alignment while surpassing the model’s native answer limits); and (iii) \emph{rich diversity} (broad HSS coverage with content diversity inherited from web sources and varied personas). 

In our experiments, we fairly compare 14 leading data-synthesis baselines across 16 mainstream benchmarks spanning eight core LLM capabilities.
Qwen3-8B-Base fine-tuned on HSS-Synth surpasses all strong baselines and sets a new SOTA record, with overall performance closest to the official instruction-tuned Qwen3-8B.
Using high-quality, diverse HSS data, HSS-Synth markedly improves human-preference alignment and, to a certain extent, enhances knowledge capabilities, effectively avoiding the ``performance seesaw'' (knowledge capabilities gains accompanied by human preference drops).
We conduct ablation studies to show HSS-Synth’s core components are indispensable; HSS-Synth transfers well across model architectures and sizes (Llama3.1-8B, Qwen2.5-14B); and different LLM capabilities require different synthetic data scales for convergence.
In-depth analysis of the HSS-Synth dataset highlights its advantages in sentence length, lexical richness, and semantic diversity.
Comparing three answer types or instructions (seed document, reverse, teacher-forced) shows that teacher-forced is the optimal type; seed documents are unsuitable as answers; and reverse answers underperform due to reuse and insufficient grounding.
Finally, we cross-validate the introduced quality rubric using both human judgments and proprietary LLM evaluations, and empirically demonstrate the necessity of text refinement.
Our contributions are as follows:
\begin{itemize}[leftmargin=8pt,itemsep=0pt,topsep=2pt]
\item HSS-Synth is the first to define a 14-field HSS domain system via a subject-centered paradigm and perform data synthesis across these domains.
\item The techniques are transferable across model architectures and broader data-synthesis tasks.
\item HSS-Synth outperforms 14 strong baselines and sets new SOTA on 16 benchmarks, improving both human preference and knowledge capabilities without a ``performance seesaw.''
\end{itemize}

\section{Related Work}
\paragraph{Humanities and Social Sciences for LLM} are central to human society and knowledge, so a strong HSS capability is essential for deploying LLMs. Yet HSS has been underemphasized in LLM research: although ~\citet{ValueByteAI_AwesomeLLM_2024} compiles a reading list for “LLM in Social Science”, it omits data synthesis. As typical open-domain tasks, HSS problems lack verifiable answers and often require nuanced human judgment, making data collection and synthesis difficult. Despite existing efforts, these work remains fragmented and limited to a few areas: \emph{writing} (simple story generation~\cite{eldan2023tinystories}, \emph{creative writing}~\cite{wang2024weaver}); \emph{dialogue systems} (task-oriented dialogue for e-commerce~\cite{qian2025bottom}, few-shot dialogue summarization~\cite{lu2025mutual}, multi-turn multi-topic dialogue~\cite{lee2025doctalk}); \emph{role-playing} (billion-scale persona construction~\cite{ge2024scaling}); and \emph{long context} (ultra-long content generation~\cite{bai2024longwriter}). 
To address this limitation, unlike the above ability-centric work, we adopt a subject-centric, application-oriented way: drawing 14 mainstream HSS domains from an official taxonomy, we build a comprehensive HSS taxonomy for LLMs and, for the first time, systematically study data synthesis in these domains.

\paragraph{Synthesizing Instruction Tuning Data} 
has evolved and falls into three categories: \emph{(1) Human-crafted methods}, e.g., WildChat~\cite{zhao2024wildchat}, which generate multi-turn human--GPT conversations; \emph{(2) Semi-automated methods}, including Evol-Instruct~\cite{xu2024wizardlm} and SynthQuestions~\cite{zhu2025real}, which expand manually annotated instructions via few-shot prompting; 
Nemotron-CC-HQ~\cite{su2024nemotron} that extracts knowledge and QA pairs; Bonito~\cite{nayak2024learning} that remixes task templates to create meta-templates for conditional generation; LongWriter~\cite{bai2024longwriter}, which uses agent-based planning for writing; 
Wrap~\cite{maini-etal-2024-rephrasing} and MGACorpus~\cite{hao2025reformulation}, both rephrasing documents, with the latter focusing on genre and audience; 
LongForm~\cite{koksal2023longform} and Back-translation~\cite{li2023self} that leverage instruction inversion; and WebR~\cite{maini-etal-2024-rephrasing}, which combines document rephrasing and instruction inversion; \emph{(3) Fully automated methods}, e.g., Magpie~\cite{xu2024magpie}, which prompts LLMs with chat templates; Cosmopedia-v2~\cite{benallal2024cosmopedia}, which conducts web-based rewriting on specific topics; and Condor~\cite{cao2025condor}, which expands world knowledge through self-refinement. 
Additionally, \emph{mixed datasets}, e.g., OpenHermes 2.5~\cite{OpenHermes} aggregate multiple open-source sources.
Building on these advances, our semi-automated HSS-Synth incorporates novel mechanisms such as QA consistency checking and teacher-forced answering for HSS data synthesis.

\vspace{-2pt}
\section{HSS-Synth}
\vspace{-4pt}
This section presents HSS-Synth in Figure~\ref{fig:method}, a three-stage pipeline for generating high-quality, diverse HSS datasets, consisting of: (1) seed document construction, (2) question-answer pair refinement, and (3) teacher-forced answering.

\subsection{Seed Document Construction}~\label{seed_doc_construct}
Given the abundant scale and diversity of web corpora~\cite{yue2024mammoth2,jiang2025instruction}, we construct seed documents from the 627B-token Slimpajama corpus~\cite{cerebras2023slimpajama}.
To address noisy web data and the scarcity of ideal HSS texts, we adopt a multi-step workflow:
We begin with \textbf{source sampling}, exclude HSS-irrelevant sources (StackExchange, GitHub), retain all documents from high-quality sources (Books, ArXiv, and Wikipedia), and randomly sample 10\% from web-crawled sources (C4, CommonCrawl) to balance source composition. Next, we apply \textbf{heuristic filtering} via the \texttt{FineWeb} toolkit\footnote{\url{https://github.com/huggingface/datatrove/blob/main/examples/fineweb.py}}, integrating fastText language classifier~\cite{joulin2016fasttext}, Gopher/C4/FineWeb quality filters~\cite{rae2021scaling,raffel2020exploring,penedo2024fineweb}, and MinHash deduplication~\cite{broder1997resemblance} to eliminate noisy and redundant documents. 
Then, we use an LLM for \textbf{domain classification} to identify texts that fall within HSS domains. 
Subsequently, we assess HSS texts using 12 expert-crafted quality rubrics organized into three tiers: \emph{readability} (grammar, coherence, content accuracy, domain relevance), \emph{applicability} (tone \& expression, knowledge depth, vocabulary richness, genre focus), and \emph{human touch} (thematic depth, emotionality, literary diversity, humanities creativity). 
The validity of these rubrics is empirically supported in Section~\ref{sec:rubric_validation}.
We prompt the LLM for \textbf{quality rating} and retain documents with readability$=5$, applicability$\geq4$, and human touch$\geq3$. 
However, residual noise, redundancy, or weak expression may remain, so we further \textbf{refine texts} with an LLM to achieve: i) content cleaning (removing crawl artifacts and irrelevant redundancy); ii) content fidelity (preserving core information, structure, and ``human-touch''); iii) expression optimization (improving coherence, fluency, and accuracy). 
Finally, a \textbf{refinement judge} verifies that refined documents satisfy these objectives and are suitable as clean seed documents. 
See Appendix~\ref{sec:appendix_details_seed_construction} for details.

\vspace{-5pt}
\subsection{Question-Answer Pair Refinement}
\paragraph{Multi-attribute Instruction Backtranslation}
Unlike prior instruction backtranslation work~\cite{koksal2023longform,li2024selfalignment}, we generate instructions from seed documents with attached multiple attributes. We first specify the ``what'' of the task: concrete, actionable requirements that guide the LLM to generate instructions that faithfully reproduce the seed document. Specifically, the instruction must (i) state the document’s domain, genre, and length; (ii) summarize the document’s core content and key points; (iii) describe the document’s structure and narrative voice; and (iv) remain concise, non-redundant, and avoid direct references such as ``source document''. We then define the ``who'' of the task—the persona the instruction should adopt to enhance diversity. To this end, we introduce persona settings~\cite{ge2024scaling} that specify stance, mindset, and tone, and uniformly use second-person phrasing. Notably, personas are co-generated with the instructions, requiring no predefined setup and thus offering greater flexibility. This multi-attribute backtranslation enables us to produce instructions that are faithful to the original documents while diverse in content.

\vspace{-2pt}
\paragraph{Question-Answer Alignment Check}
\vspace{-4pt}
Although instruction backtranslation is effective, it may introduce alignment bias—i.e., the reverse instruction (\textit{question}) is misaligned with the seed document (\textit{seed answer})—which violates the strict Q\&A alignment required for instruction tuning, a critical issue overlooked by prior work. To address this, we propose a Q\&A alignment check: an LLM first answers the reverse instruction to obtain a \textit{reverse answer}, then we measure the consistency between the seed answer and the reverse answer to verify whether the reverse instruction can faithfully reproduce the seed document. The checker yields binary decisions with rationales along two dimensions: (1) textual consistency—alignment of core content and key points; and (2) instruction effectiveness—covering essential information without explicit references. We retain only Q\&A pairs that pass both checks, further improving the fidelity of reverse instructions.

\vspace{-4pt}
\subsection{Teacher-Forced Answering}
\vspace{-5pt}
Even after question-answer pair refinement yields usable reversed instruction–answer pairs, reverse answers generated solely by an LLM may exhibit factual gaps, sparse detail, and style drift~\cite{cao2025condor,jiang2025instruction}. Hence, we propose teacher-forced answering (TeachForceA), inspired by teacher forcing—feeding external ground-truth inputs into the model at each step~\cite{lamb2016professor}. Specifically, when answering an instruction, the seed document (i.e., information source) is fed into the LLM alongside the instruction as a semantic anchor. The seed document need not outperform the reverse answer; it only needs to (1) cover the information, details, and examples required by the instruction; (2) convey the original tone and style; and (3) supply key vocabulary and sentence structures in certain passages. We also explicitly require the LLM that, in case of conflict, the faithful answer to the instruction prevails, preserving strict Q\&A consistency. By confining information integration to the controlled set "reversed instruction + seed document," TeachForceA (i) exploits undistorted facts from the seed document to reduce hallucinations; (ii) transfers the original tone and style to enhance human touch; and (iii) uses structural and lexical cues to improve completeness and readability—thereby surpassing the ceiling of traditional LLM-distilled answers.

\begin{table*}[t]
  \footnotesize
  \centering
  \renewcommand{\arraystretch}{1.1}
  \setlength{\tabcolsep}{1.5pt}{
  \resizebox{1.\textwidth}{!}{
  \begin{tabular}{llccccccccc}
  
    \toprule

    \multicolumn{1}{l}{\multirow{2}{*}{\textbf{Selected Method}}}   & 
    \multicolumn{1}{l}{\multirow{2}{*}{\textbf{Synthesis Model}}}   & 
    \multicolumn{3}{c}{\multirow{1}{*}{\textbf{\begin{tabular}[c]{@{}c@{}}Human Preference \end{tabular}}}} & 
    \multicolumn{5}{c}{\multirow{1}{*}{\textbf{\begin{tabular}[c]{@{}c@{}}Knowledge based \end{tabular}}}} &
    \multicolumn{1}{c}{\multirow{2}{*}{\textbf{AVG}}} \\
    
    \cmidrule(lr){3-5}\cmidrule(lr){6-10}

    \multicolumn{2}{l}{}   & \textbf{Writing}   & \textbf{Emotion} & \textbf{Social} & \textbf{Instruct}  & \textbf{World} & \textbf{Commonsense}  & \textbf{LongCtx} & \textbf{Comprehension} \\

    \midrule
        Qwen3-8B~\cite{yang2025qwen3} & \multicolumn{1}{l}{--} & 64.52 & 49.20 & 66.63 & 56.13 & 74.90 & 76.49 & 47.65 & 35.22 & 60.70 \\
        Qwen3-8B-Base~\cite{yang2025qwen3} & \multicolumn{1}{l}{--}  & 31.95 & 25.10 & 61.91 & 31.18 & 77.07 & 76.89 &46.24 &32.76 & 50.73\\
        \midrule
        WildChat~\cite{zhao2024wildchat} &  ChatGPT &32.37 & 40.52  & 63.25 & 38.96& 77.01 & 79.70 & 45.50 & 35.97 &51.66  \\
        OpenHermes 2.5~\cite{OpenHermes} &  Mixed &41.04 & 38.24& 69.16 & 35.77 & 77.16 & 79.02  & 50.93  &30.71  & 52.75\\
        SynthQuestions~\cite{zhu2025real} & Llama3-70B-Instruct &40.55&37.34  & 68.63 & 35.08  & 77.01 & 79.04 &51.29 & \textbf{37.90} &53.36 \\
        Bonito~\cite{nayak2024learning} &  Mistral-7B & 22.89 & 3.42 & \textbf{72.94} & 20.05 & 75.80  & 79.05 & 45.76 & 34.89   &44.35  \\
        Evol Instruct~\cite{xu2024wizardlm} & ChatGPT & 29.38 & 31.86  & 62.40 & 39.16 & 77.28 & 80.21 & 49.91 &37.48 &50.96  \\
        LongWriter~\cite{bai2024longwriter} & GPT-4o & 39.82 & 39.91 & 38.78 & 27.44 & 77.01 & \textbf{80.39} & 44.07 & 15.83    &45.41 \\
        Magpie~\cite{xu2024magpie} & Mixed & 42.59 & 40.17 & 68.54 & 42.90 & \textbf{77.50}  & 80.06  & 48.74 & 36.3 &54.59 \\
        Nemotron-CC-HQ~\cite{su2024nemotron} & Mistral-NeMo-12B-Ins. &29.29 & 28.03  & 58.98 & 31.40  & 76.54 & 79.95 & 48.72 & 26.19 &47.39 \\ 
        MGACorpus~\cite{hao2025reformulation} & 3.3B MoE &36.02 & 36.04 & 66.43 & 36.75 & 76.89 & 77.91 & \textbf{51.51} & 31.79 &51.67 \\
        Cosmopedia-v2~\cite{benallal2024cosmopedia} & Mixtral-8x7B-Instruct & 36.61 & 30.67 & 65.39 & 35.10 & 77.41 & 79.95 & 48.45 & 32.84 &50.71 \\
        Condor~\cite{cao2025condor} & Qwen2.5-72B & 39.49 & 44.66  & 65.73 & 27.31 & 77.28 & 80.07 & 43.31 & 24.16 &50.25 \\
        WebR-Pro~\cite{jiang2025instruction} & GPT-4o-mini & 49.50 & 45.36  & 66.55  & 43.25 & 77.09 & 79.93 & 46.35 & 35.92 &55.49 \\ 
        WRAP~\cite{maini-etal-2024-rephrasing} & Qwen-30B-A3B & 21.45 & 14.63 & 55.90 & 19.90 & 76.74 & 78.53 & 45.43 & 27.35 &42.50 \\
        LongForm~\cite{koksal2023longform} & Qwen-30B-A3B & 21.18 & 6.30 & 35.29 & 10.89 & 76.85 & 79.23 & 40.72 & 30.13 &37.57 \\
        \multicolumn{11}{l}{\textbf{HSS-Synth (Ours)}} \\
        \rowcolor[gray]{0.97}\hspace{.8em}-w/ MA-IBT & Qwen3-30B-A3B  & 56.71 & 46.02  & 58.97  & 38.67  & 77.17 & 78.37 & 28.37 & 25.83 &51.26\\ 

        \rowcolor[gray]{0.97}\hspace{.8em}-w/ MA-IBT \& QAC & Qwen3-30B-A3B  & 58.86 & 51.28  & 65.86 & 38.67 & 77.24 & 77.77 & 47.24 & 24.86 &55.22\\ 
      
        \rowcolor[gray]{0.97}\hspace{.8em}-w/ MA-IBT \& QAC \& TeachForcedA & Qwen3-30B-A3B  & \textbf{60.01} & \textbf{51.31} & 67.67  & \textbf{47.70}  & 76.81 & 77.86 & 48.71 & 32.62 & \textbf{57.84} \\ 
    \bottomrule
  \end{tabular}
  }}
\vspace{-5pt}
\caption{
16 Benchmark results covering 8 core capabilities for Qwen3-8B-Base trained on instruction-tuning data synthesized by different methods. HSS-synth surpasses all baselines (SOTA), nearly matches the official instruct-tuned Qwen3-8B, and improves both human preference and knowledge without performance trade-offs.
Here, the abbreviation MA-IBT, QAC denotes multi-attribute instruction backtranslation, Q\&A alignment check.
}
  \label{tab:main_result}
  \vspace{-12pt}
\end{table*}

\vspace{-3pt}
\section{Experiment}
\vspace{-4pt}
\subsection{Experimental Setup}
\vspace{-3pt}
\paragraph{Baselines}
We compare the instruction tuning (IT) dataset generated by HSS-Synth with 14 leading open-source IT datasets, all containing 230k instruction--response pairs\footnote{We use all data if a dataset has fewer than 230k pairs.}, grouped as follows: (1) Human-crafted, e.g., \textbf{WildChat}~\cite{zhao2024wildchat}; (2) Mixed data, such as \textbf{OpenHermes 2.5}~\cite{OpenHermes}; (3) Semi-automated synthetic data, including \textbf{Bonito}~\cite{nayak2024learning}, \textbf{Evol-Instruct}~\cite{xu2024wizardlm}, \textbf{Nemotron-CC-HQ}~\cite{su2024nemotron}, \textbf{LongWriter}~\cite{bai2024longwriter}, \textbf{MGACorpus}~\cite{hao2025reformulation}, \textbf{WebR}~\cite{jiang2025instruction}, and
\textbf{SynthQuestions}~\cite{zhu2025real} (with LongWriter focusing on writing tasks); (4) Fully automated synthetic data, such as \textbf{Magpie}~\cite{xu2024magpie}, \textbf{Cosmopedia-v2}~\cite{benallal2024cosmopedia}, and \textbf{Condor}~\cite{cao2025condor}. To ensure fair comparison, we reproduce two representative methods---document rephrasing (\textbf{Wrap}~\cite{maini-etal-2024-rephrasing}) and instruction inversion without persona (\textbf{LongForm}~\cite{koksal2023longform})---using the same seed data and synthesis models as ours.

\vspace{-5pt}
\paragraph{Data Synthesis Settings}
Throughout our data synthesis pipeline from large-scale web corpora, we consistently used Qwen3-30B-A3B~\cite{yang2025qwen3}, chosen because it offers both strong performance (see Table~\ref{tab:main_result}) and cost efficiency (deployable on a single H800 GPU), with the default generation configuration (temperature=0.6, top-k=20, top-p=0.95). After filtering out improper JSON formats and removing \texttt{<thinking>} content, we obtained 230k instruction–answer pairs for fine-tuning. The data synthesis process consumed 2,400 GPU hours.

\begin{table*}[t]
  \footnotesize
  \centering
  \renewcommand{\arraystretch}{1.1}
  \setlength{\tabcolsep}{2pt}{
  \resizebox{1.\textwidth}{!}{
  \begin{tabular}{lcccccccccc}
  
    \toprule

    \multicolumn{1}{l}{\multirow{2}{*}{\textbf{Selected Method}}}   & 
    \multicolumn{3}{c}{\multirow{1}{*}{\textbf{\begin{tabular}[c]{@{}c@{}}Human Preference \end{tabular}}}} & 
    \multicolumn{5}{c}{\multirow{1}{*}{\textbf{\begin{tabular}[c]{@{}c@{}}Knowledge based \end{tabular}}}} &
    \multicolumn{1}{c}{\multirow{2}{*}{\textbf{AVG}}} \\
    
    \cmidrule(lr){2-4}\cmidrule(lr){5-9}
    
    \multicolumn{1}{l}{}   & \textbf{Writing}   & \textbf{Emotion} & \textbf{Social} & \textbf{Instruct}  & \textbf{World} & \textbf{Commonsense}  & \textbf{LongCtx} & \textbf{Comprehension} \\

    \midrule
        Qwen2.5-14B~\cite{qwen2.5}  & 23.94 & 31.17 & \textbf{68.87} & 32.26 & \textbf{80.45} & \textbf{82.06} & \textbf{51.33} & \textbf{37.46} & 49.70 \\
        Wrap~\cite{maini-etal-2024-rephrasing} & 20.60 & 22.58 & 62.76 & 19.32 &80.06 & 81.97 & 46.36 & 19.20 & 44.11 \\
        LongForm~\cite{koksal2023longform} &  15.09 & 9.49 &  60.61 & 25.30 & 79.89 & 80.22 & 47.30 & 28.91 & 43.35 \\
        \rowcolor[gray]{0.97} \textbf{HSS-Synth (Ours)}  & \textbf{57.50} & \textbf{50.95} & 65.59 & \textbf{46.70} & 80.22 & 80.79 & 49.73 & 33.52 & \textbf{57.00}\\ 
    \midrule
    \midrule
        Llama3.1-8B~\cite{llama3modelcard}  & 9.17 & 8.41 & 35.91 & 17.95 & \textbf{66.93} & 79.97 & 43.11 &14.04 & 34.44 \\
        Wrap~\cite{maini-etal-2024-rephrasing} & 8.14 & 3.47 & 38.11 & 6.33 & 65.89 & \textbf{80.86} & 37.72 & 27.02 & 32.19 \\
        LongForm~\cite{koksal2023longform}  & 18.46 & 5.58 & 41.71 &11.81 & 66.77 & 80.40 & 40.86 & \textbf{30.45} & 36.95 \\
        \rowcolor[gray]{0.97} \textbf{HSS-Synth (Ours)}  & \textbf{39.95} & \textbf{46.45}  & \textbf{50.46} & \textbf{24.71}  & 66.14 & 79.26 & \textbf{46.63} & 28.37 & \textbf{47.75} \\ 
    \bottomrule
  \end{tabular}}
  }
 \vspace{-4pt}
  \caption{
Performance comparison for different methods across model architectures (Llama‑3.1‑8B) and scales (Qwen2.5‑14B), where HSS‑Synth achieves the best cross‑model results.
  }
  \label{tab:differ_model}
   \vspace{-10pt}
\end{table*}

\vspace{-5pt}
\paragraph{Model Training Settings}
For instruction tuning, we trained Qwen3-8B-base, Qwen2.5-14B-base~\cite{qwen2.5}, and LLaMA3.1-8B-base~\cite{llama3modelcard} on various datasets, using the same hyperparameters for fair comparison. Full training details are provided in Appendix~\ref{sec:appendix_training}.

\vspace{-5pt}
\paragraph{Evaluation Benchmarks and Metrics}~\label{sec:benchmark_metric}
We evaluate fine-tuned models on 16 mainstream benchmarks spanning 8 core LLM capabilities, grouped into \emph{human-preference} and \emph{knowledge-based} tasks. 
For each benchmark, we report results under its default setting (0-shot unless specified), along with the metric and judge model when applicable.
\textbf{Writing Skill}: WritingBench (rubric score, Claude-3.7-Sonnet)~\cite{wu2025writingbench}, CreativeWriting-v3 (rubric score, Claude-3.7-Sonnet), and Judgemark-v2 (Judgemark score)~\cite{paech2023eq};
\textbf{Emotional Perception}: BuzzBench (rubric score, Claude-3.5-Sonnet-v2) and EQ-Bench3 (rubric score, Claude-3.7-Sonnet)~\cite{eqbench3_repo_2025};
\textbf{Social Interaction}: Social-IQA (accuracy)~\cite{sap2019socialiqa}, IQuiz\_EQ (accuracy)~\cite{chen2024tombench};
\textbf{Instruction Following}: IFEval (prompt-level strict accuracy)~\cite{zhou2023instruction}, Collie (aggregate accuracy)~\cite{yao2023collie};
\textbf{World Knowledge}: MMLU (5-shot, accuracy; we also report result on the HSS subset)~\cite{hendrycks2020measuring};
\textbf{Commonsense Reasoning}: HellaSwag (normalized accuracy)~\cite{zellers2019hellaswag}, StoryCloze (accuracy)~\cite{mostafazadeh2016corpus};
\textbf{Long-Context}: CoQA (F1)~\cite{reddy2019coqa}, GovReport (ROUGE-L)~\cite{huang2021efficient};
\textbf{Reading Comprehension}: NarrativeQA (accuracy, Llama-3.3-70B-Instruct)~\cite{kovcisky2018narrativeqa}, XSum (ROUGE-L)~\cite{narayan2018don}.
To ensure reliable evaluation, we perform a standard 13-gram decontamination analysis and find zero overlap between the synthesized data and the benchmark test sets.

\vspace{-5pt}
\subsection{Main Results}
\vspace{-3pt}
We report benchmark results of Qwen3-8B-Base fine-tuned on different instruction-tuning datasets in Table~\ref{tab:main_result}, from which we draw the insights below:

\begin{itemize}[leftmargin=0pt,label={},itemsep=-2pt,topsep=2pt]
\item \textbf{Synthetic baselines are strong}
Most synthetic-data baselines outperform the base model. Notably, WebR-Pro—synthesized from web corpora—stands out, validating web-based synthesis as a promising path, consistent with the design intent of HSS-Synth.

\item \textbf{Synthesize model capability and data domain coverage matter}
Bonito relies solely on Mistral-7B for synthesis and underperforms, confirming that data quality hinges on the synthesis model’s capability. In contrast, using only Qwen3-30B-A3B as our synthesis model already matches many strong baselines, evidencing the effectiveness of HSS-Synth. Moreover, LongWriter focuses narrowly on writing-domain data and is therefore limited to most tasks, whereas HSS-Synth benefits from broad coverage across HSS domains.

\item \textbf{Advanced text refinement and multi-attribute instruction backtranslation}
Under the same settings, WRAP and LongForm apply only basic document rephrasing and instruction backtranslation and thus lag behind, highlighting the advantage of our stronger text refinement and multi-attribute instruction backtranslation techniques.

\item \textbf{HSS-Synth attains SOTA without a performance seesaw}
Qwen3-8B-base fine-tuned on HSS-Synth achieves the best average across 16 mainstream benchmarks and is overall closest to the official instruction-tuned Qwen3-8B. Specifically, HSS-Synth delivers large gains on human-preference evaluations; on knowledge-oriented benchmarks, improvements in instruction following are most pronounced, while other capabilities are on par with or superior to the official base/instruct models. This indicates that high-quality HSS data not only strengthens human preference but also, to some extent, enhances knowledge-intensive performance. Crucially, HSS-Synth balances preference and knowledge, effectively avoiding the ``performance seesaw,'' wherein gains on knowledge capabilities are accompanied by drops in human preference~\cite{dubois2024length}.
\end{itemize}

\vspace{-4pt}
\subsection{Ablation Study}
\vspace{-2pt}

\begin{itemize}[leftmargin=0pt,label={},itemsep=-2pt,topsep=2pt]
\item \textbf{Core Components}
Table \ref{tab:main_result} reports the performance of HSS-Synth under different component settings:
(1) \textbf{w/ MA-IBT}: compared to LongForm (i.e., without MA-IBT), multi-attribute instruction backtranslation clarifies task attributes (what, who), yielding faithful and diverse instructions, so synthetic data becomes usable and surpasses the base model.
(2) \textbf{w/ MA-IBT \& QAC}: adding Q\&A alignment checks further improves backtranslated instruction faithfulness, significantly boosting data quality to lead performance gains.
(3) \textbf{w/ MA-IBT \& QAC \& TeachForcedA}: teacher-forced answering overcomes LLM’s inherent response constraints, achieving the best performance.
Overall, these components are validated as essential and indispensable to the pipeline.

\item \textbf{Model Structures and Sizes}
Table \ref{tab:differ_model} compares, across model architectures and scales, models trained on HSS-Synth synthetic data, the official Qwen3-base, and two faithfully reproduced baselines (Wrap, LongForm). HSS-Synth consistently outperforms the official base and both baselines, with its advantage most pronounced in human-preference evaluations; on knowledge-based benchmarks, it shows the smallest gap to Qwen2.5-14B while clearly outperforming Llama-3.1-8B, demonstrating its strong cross-model applicability.

\begin{figure}[t]                         
  \centering
  \includegraphics[width=\linewidth]{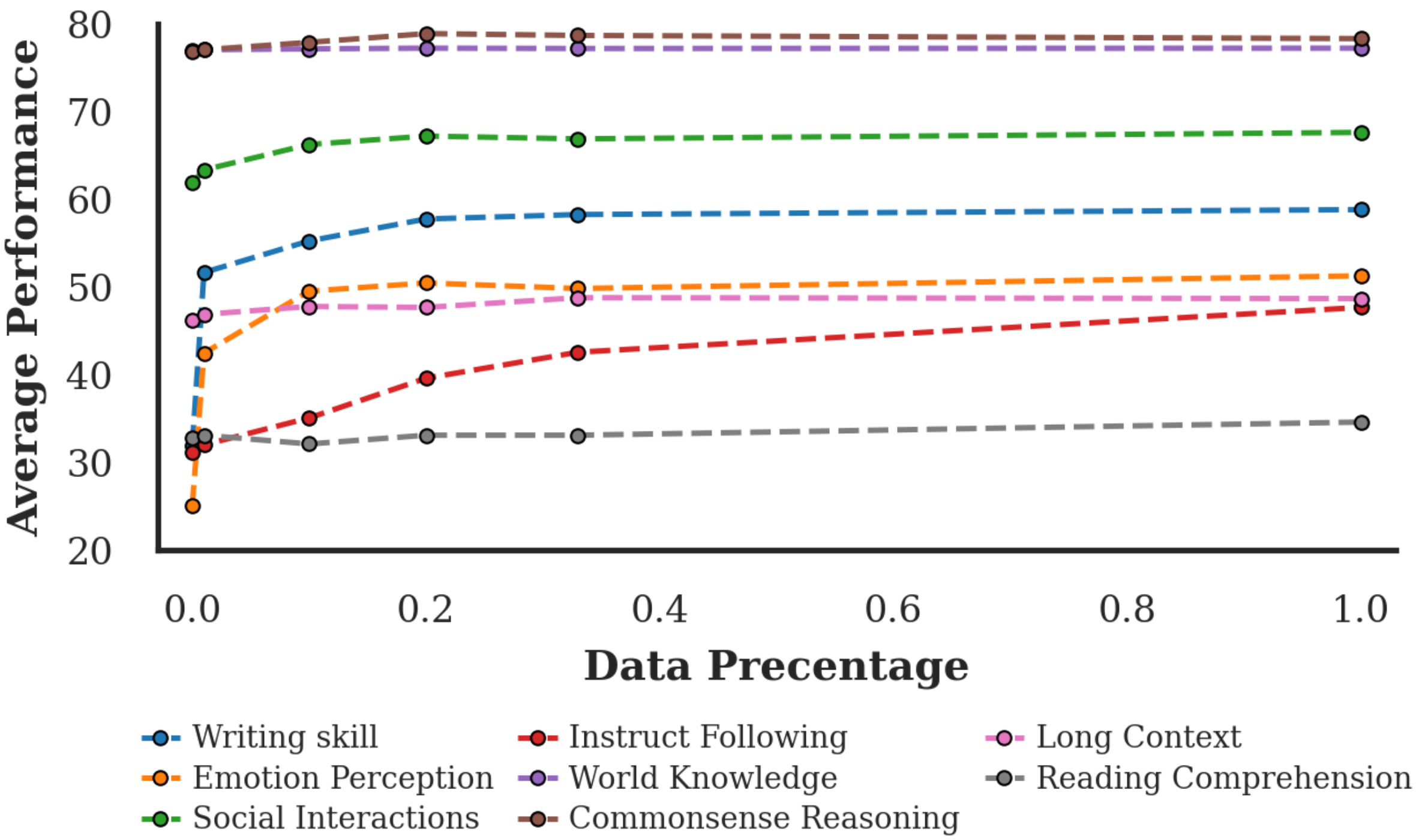}
  \caption{
Impact of HSS-Synth data scale on 8 LLM capabilities. Knowledge skills converge at $\approx$ 10\%, human-preference at $\approx$20\%, and instruction following at $\approx$33\% of the data (all within tens of thousands of samples). Notably, with no observed cross-skill trade-offs.
  }
  \label{fig:scaling_effect}
   \vspace{-5pt}
\end{figure}

\item \textbf{Synthetic Data Scales} 
Figure \ref{fig:scaling_effect} illustrates the impact of HSS-Synth data scaling on different LLM capabilities. We plot scaling curves for eight skills and estimate their convergence data requirements. As the data ratio increases, knowledge-based abilities—except instruction following—exhibit diminishing returns around 10\% ($\approx$23k samples), reflecting that SFT primarily strengthens instruction following rather than injecting new factual knowledge. Human-preference skills plateau at around 20\% ($\approx$46k samples), indicating that the base model already encodes relevant concepts and requires limited data for style alignment. Instruction following benefit from diversity up to about 33\% ($\approx$76k samples) before saturating. Overall, for strong base models like Qwen3-8B, convergence typically occurs within tens of thousands of samples, with marginal gains beyond. Notably, we observe no cross-skill trade-offs: human preference gains do not harm knowledge skills and even slightly improve them.
\end{itemize}

\vspace{-8pt}
\subsection{Data Analysis}
\vspace{-4pt}
\begin{table}[t]
\centering
\small
\setlength{\tabcolsep}{2pt}{
\resizebox{1.\linewidth}{!}{
\begin{tabular}{lccc}
\toprule
\textbf{Selected Method} & \textbf{Avg. Input Len} & \textbf{Avg. Output Len} & \textbf{MTLD} \\
\midrule
WildChat           & 400  & 422  & 70.78  \\
OpenHermes 2.5     & 136  & 214  & 53.65  \\
Nemotron\_CC\_HQ   & 3,025 & 401  & 94.82  \\
MGACorpus          & 1,330 & 779  & 92.45  \\
CosmopediaV2       & 332  & 687  & 158.40 \\
Bonito             & 2,634 & 134  & 89.64  \\
Evlo-instruct      & 123  & 355  & 56.63  \\
WebR Pro           & 464  & 460  & 82.51  \\
SynthQuestions     & 83   & 762  & 67.28  \\
LongWriter         & 282  & 4,589 & 124.52 \\
WRAP               & 172  & 487  & 62.95  \\
LongForm           & 120  & 2,696 & 108.41 \\
Magpie             & 33   & 440  & 70.63  \\
Condor             & 70   & 876  & 97.55  \\
\rowcolor[gray]{0.97} \textbf{HSS-Synth (Ours)} & \textbf{253} & \textbf{1,635} & \textbf{127.50} \\
\bottomrule
\end{tabular}
}
\vspace{-6pt}
\caption{Average input/output length and measure of textual lexical diversity (MTLD) of various instruction–tuning datasets.}
\vspace{-5pt}
\label{tab:dataset_stats}}
\end{table}

\paragraph{Data Length and Lexical Richness} 
Table~\ref{tab:dataset_stats} summarizes average input/output token counts and measures of textual lexical diversity (MTLD)~\cite{mccarthy2010mtld} for instruction-tuning datasets. Tokens are counted with the Qwen3 tokenizer. MTLD equals total tokens divided by the number of factors, where a factor forms when the type/token ratio (TTR) falls below a threshold; higher MTLD indicates greater lexical diversity. HSS‑Synth back-translates instructions from web-corpus seed documents, yielding input lengths comparable to web-based WebR‑Pro and WildChat. Benefiting from TeachForcedA, HSS‑Synth averages over 1.6k tokens per answer, meeting the length needs of HSS domains. Encouragingly, while providing long outputs, HSS-Synth attains the second-highest MTLD, only behind the synthetic dataset Cosmopedia2 designed for pre-training, highlighting the rich lexical signals it offers for model fine-tuning.

\begin{figure}[t]
\centering
\begin{subfigure}{0.65\linewidth}
\centering
\includegraphics[width=\linewidth]{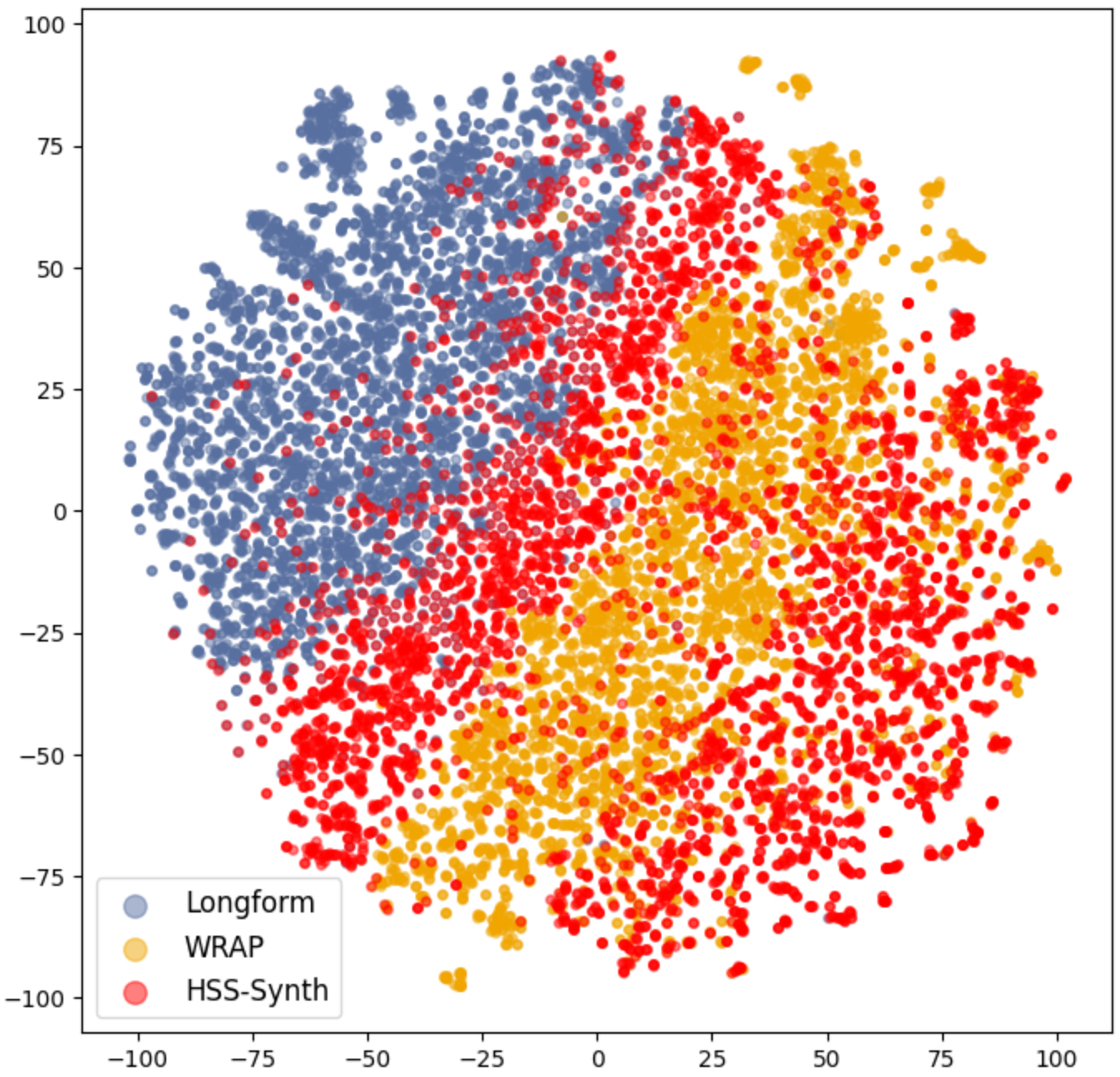}
\label{fig:tsne_embedding}
\end{subfigure}\hfill
\begin{subfigure}{0.35\linewidth}
\centering
\includegraphics[width=\linewidth]{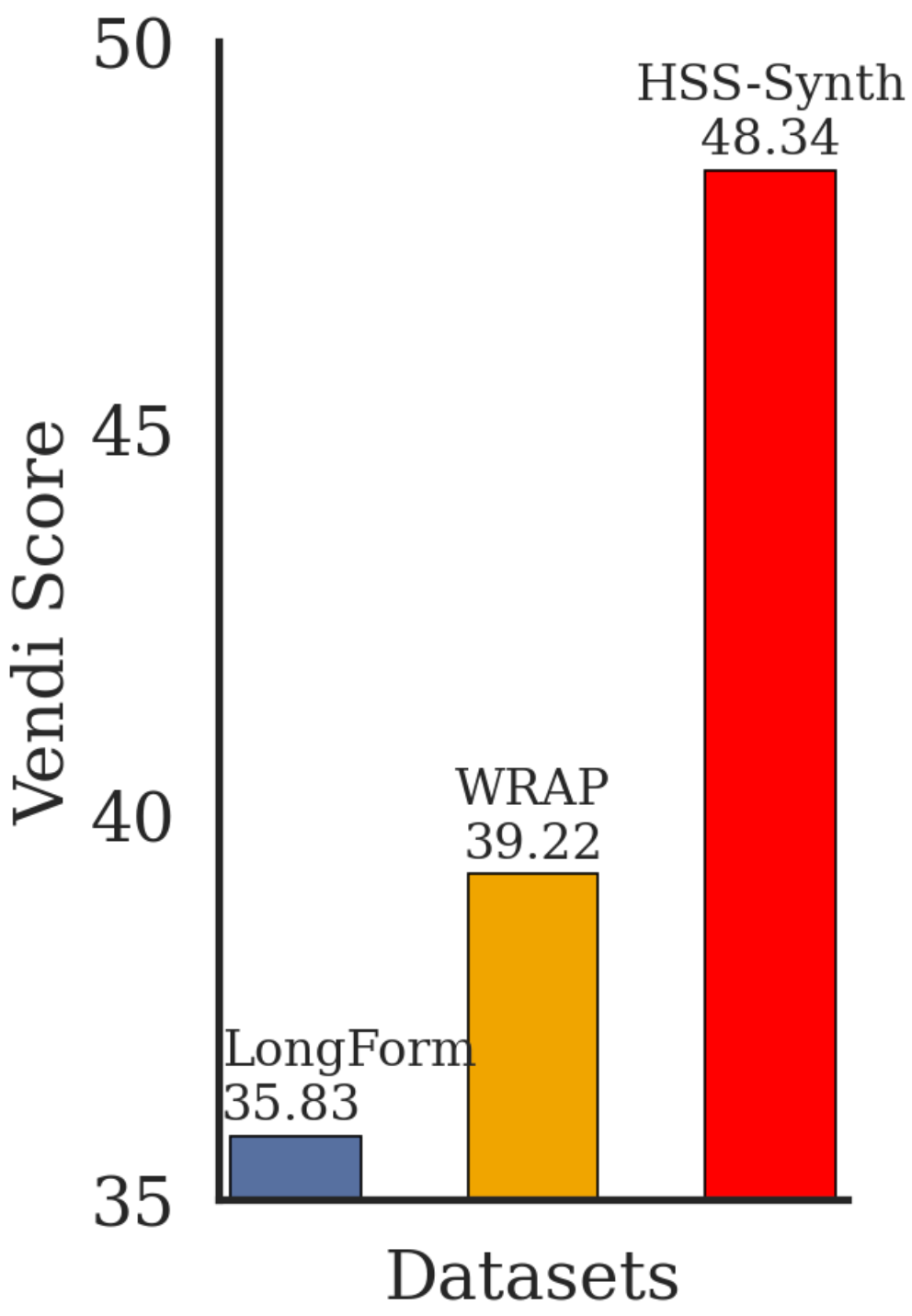}
\label{fig:vendi_score}
\end{subfigure}
\vspace{-28pt}
\caption{t-SNE visualization (left) and Vendi scores (right) of sentence embeddings for LongForm, HSS-Synth, and WRAP.}
\label{fig:tsne_embedding_vendi}
\vspace{-18pt}
\end{figure}

\vspace{-6pt}
\paragraph{Semantic diversity of Dataset}
We assess the semantic diversity of instruction–answer pairs produced by three data synthesis methods. From each dataset, we randomly sample 10k instances, feed them into gte‑Qwen2‑7B‑instruct~\cite{li2023towards} to compute sentence embeddings, and project them to a 2D space using t‑SNE. In the left panel of Figure~\ref{fig:tsne_embedding_vendi}, HSS‑Synth exhibits a wider spread, while LongForm and WRAP are more clustered. We also report the Vendi Score~\cite{dan2023vendi}, which constructs a Gram matrix from embedding similarities and measures diversity via the von Neumann entropy of its eigenvalues. In the right panel of Figure~\ref{fig:tsne_embedding_vendi}, HSS‑Synth attains the highest Vendi Score among all synthetic datasets. Both quantitative and qualitative results indicate that, under the same seeds and synthesis model, HSS‑Synth produces more semantically diverse data, providing richer training signals and thereby improving model generalization.

\begin{table}[t]
\centering
\setlength{\tabcolsep}{0.5mm}
\resizebox{0.9\linewidth}{!}{
\begin{tabular}{lccc}
\toprule
 & \textbf{Rubric Score} & \textbf{Avg. Sent Len} & \textbf{MTLD} \\
\midrule
Seed Doc & 50.75 & 2,144 & 114.56 \\
Reverse A & 54.29 & 2,621 & 123.50 \\
TeachForced A & 54.89 & 1,660 & 127.93 \\
\bottomrule
\end{tabular}
}
\vspace{-4pt}
\caption{Quality rubric score, average sentence length, and MTLD for three answers to the reverse instruction.}
\label{tab:teacher_forced_stats}
\vspace{-6pt}
\end{table}

\begin{figure}[t] 
    \centering
    \includegraphics[width=0.7\linewidth]{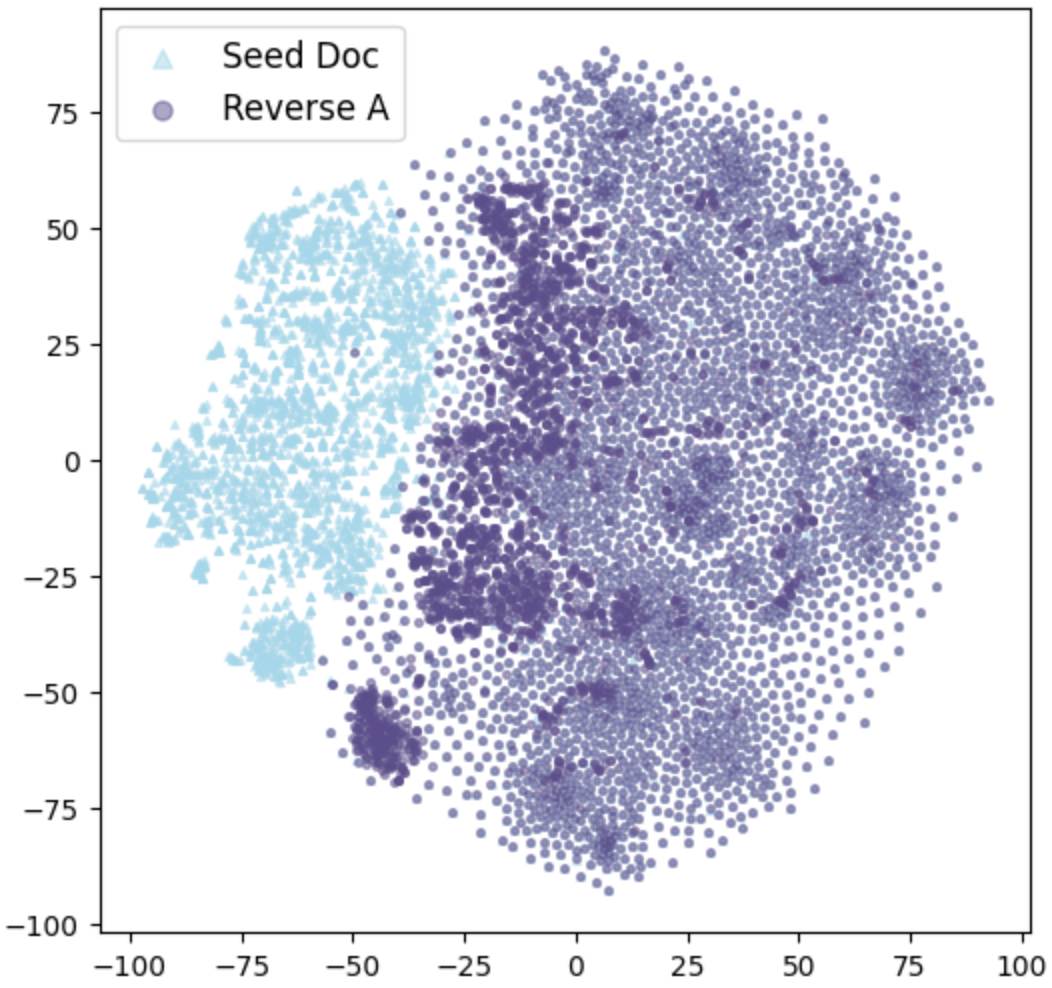}
     \vspace{-4pt}
    \caption{t-SNE plot of sentence embeddings for reverse answers and seed documents.}
    \label{fig:tsne-seed-inverted}
    \vspace{-17pt}
\end{figure}

\begin{table}[t]
\centering
\setlength{\tabcolsep}{0.5mm}
\resizebox{1.\linewidth}{!}{
\begin{tabular}{lccc}
\toprule
& \textbf{4-gram Overlap} & \textbf{LCS Length} & \textbf{Copy Ratio} \\
\midrule
Reverse A & 25.11 & 31.35 & 0.42 \\
TeachForced A & 27.35 & 64.77 & 0.78 \\
\bottomrule
\end{tabular}
}
\vspace{-4pt}
\caption{Reuse metrics from seed documents for reverse and teacher-forced answers.}
\label{tab:reuse_metrics}
\vspace{-18pt}
\end{table}

\vspace{-4pt}
\subsection{Evaluation of Different Answers}
\vspace{-2pt}

\paragraph{Rubric scores, sentence length, lexical diversity, and semantic embeddings}
Table \ref{tab:teacher_forced_stats} reports quality rubric scores, average sentence length, and measure of textual lexical diversity (MTLD) for three types of answers to the same instruction:

1) Teacher-forced answers achieve the highest rubric score and MTLD while being the shortest sentence length, indicating the best performance with the most concise phrasing. Together with component ablations, this further supports teacher-forced as the optimal answer type for instructions.

2) Reverse answers are used as the reference baseline because they are generated directly for the instruction by the synthesis model, inherently ensuring Q\&A consistency. They also show high rubric and MTLD scores and perform well under the w/ MA-IBT \& QAC ablations.

3) Seed documents are not selected as answers for two reasons: 
(i) lower quality—their rubric and MTLD scores are clearly below the other two; 
(ii) distribution shift—t-SNE in Figure \ref{fig:tsne-seed-inverted} shows sentence embeddings of seed documents (blue) and reverse answers (red) forming two distinct clusters with an average cosine similarity of only 0.20±0.05, indicating substantial divergence. 
This indicates a distribution shift between seed documents and reverse answers, and using seed documents as answers may induce Q\&A mismatch; instead, they are better suited as factual sources, consistent with recent findings \cite{zhu2025real,jiang2025instruction}.

\vspace{-5pt}
\paragraph{Inference log-likelihood heatmaps}
We train two models with (instruction, seed document) and (instruction, reverse answer) pairs, and visualize inference token-level log-likelihood on the validation set in the Appendix's heatmaps \ref{fig:seed_logprob_heatmap},\ref{fig:reverse_logprob_heatmap}. Darker colour denotes tokens that the model fails to predict confidently. The model trained on reverse answers is predominantly green, suggesting the answer pattern is easy to learn, whereas the model trained on seed answers exhibits frequent red tokens, indicating harder learning and potential errors.

\vspace{-5pt}
\paragraph{Reuse metrics from seed documents}
To quantify how much reverse and teacher-forced answers are extracted from seed documents, we randomly sampled 10k instances and computed three reuse metrics: i) average 4-gram overlap, ii)- longest common substring (LCS) length, iii)-length-normalised copy ratio = 2 × LCS / $|$answer$|$. As shown in Table 4, teacher-forced answers outperform reverse answers on all token- and substring-level reuse metrics, indicating they anchor on seed documents to extract original statements rather than generating answers from scratch like reverse answers. This strong textual grounding also explains its superior performance.

\vspace{-5pt}
\subsection{Validating Quality Rubric Effectiveness}~\label{sec:rubric_validation}
Here we validated the expert-crafted HSS quality rubric on 400+ randomized seed texts spanning diverse sources, lengths, domains, and quality tiers. Twenty uninvolved raters independently scored each item on a five-point scale, and Qwen3-30B-A3B and GPT-4.1also applied the rubric. The synthesis model matched majority-vote human labels in 86\% of cases and showed 91\% agreement with the GPT-4.1. Inter-rater agreement, summarized by Cohen’s kappa~\cite{mchugh2012interrater}, indicated strong consistency among human raters.  Overall, cross-checking human judgments and LLM outputs provides strong evidence of the rubric’s validity.

\vspace{-4pt}
\section{Conclusion}
\vspace{-4pt}
In this paper, we introduce \textbf{HSS-Synth}, the first method for synthesizing humanities and social sciences data for LLMs. Our three-stage pipeline successfully generated over 230k high-quality, diverse samples across 14 HSS domains, significantly improving model fine-tuning performance. Comprehensive evaluations, ablation studies, and in-depth analyses validate its effectiveness. This work not only alleviates the scarcity of HSS data but also advances the field of open-domain data synthesis.

\section*{Limitations}
HSS-Synth’s three-stage pipeline depends on the synthesis model and may introduce LLM bias, a common limitation of current data synthesis methods. The synthetic dataset currently covers 14 mainstream HSS domains (e.g., education, economics, law) but misses subfields such as anthropology and religious studies, and its domain distribution shows a pronounced long tail and imbalance; further work should improve domain coverage and balance. For commonsense reasoning, long-context understanding, and reading comprehension, model fine-tuning yields only marginal gains, indicating that most knowledge is injected during pre-training rather than SFT; we therefore plan to explore pre-training data synthesis and the associated scaling laws to more effectively strengthen model capabilities.

\section*{Ethical Considerations}
All authors attest that this work strictly adheres to the ACL Code of Ethics. Below, we summarize the ethical considerations specific to this study.
First, we report the limitations of this work and ensure that it does not pose any potential risks.
Second, all code, data, and models used are open source, released under the Apache-2.0 license for academic research, and contain no discrimination, personally identifiable information, or other harmful content.
Furthermore, we provide detailed reporting of our computational experiments and the procedures involving human annotators.
Finally, we used AI assistants solely to support the language of the paper—paraphrasing or polishing the authors’ original content.

\section*{Acknowledgements}
This work was mainly supported by the National Natural Science Foundation of China (NSFC) under Grant No. 62402424.

\bibliography{custom}

\clearpage
\appendix
\section*{\makebox[\linewidth][c]{\Large Appendix}}

\begin{figure*}[h]
    \centering
    \begin{subfigure}{\linewidth}
        \centering
        \includegraphics[page=1,width=1.\linewidth]{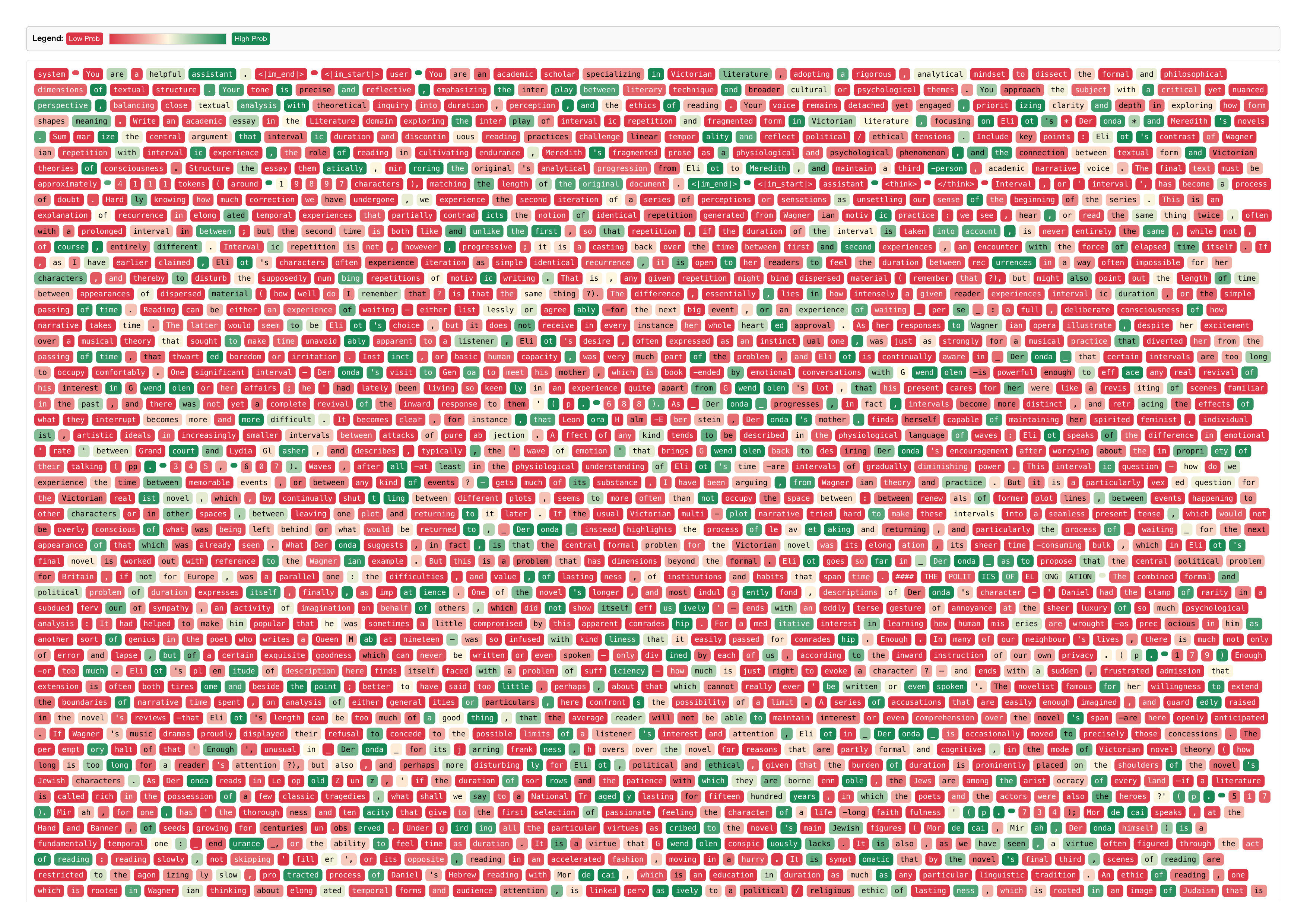}
        \vspace{-30pt} 
    \end{subfigure}
    \begin{subfigure}{\linewidth}
        \centering
        \includegraphics[page=2,width=1.\linewidth]{figures/seed_logprob_heatmap.pdf}
        \vspace{-35pt}
    \end{subfigure}
    \caption{Inference token likelihood heatmaps for models trained on \textbf{seed documents}, where darker colors indicate tokens the model fails to predict confidently. The model trained on seed answers clearly finds its answer pattern hard to learn.}
    \label{fig:seed_logprob_heatmap}
\end{figure*}

\begin{figure*}[h]
    \centering
    \begin{subfigure}{\linewidth}
        \centering
        \includegraphics[page=1,width=1.\linewidth]{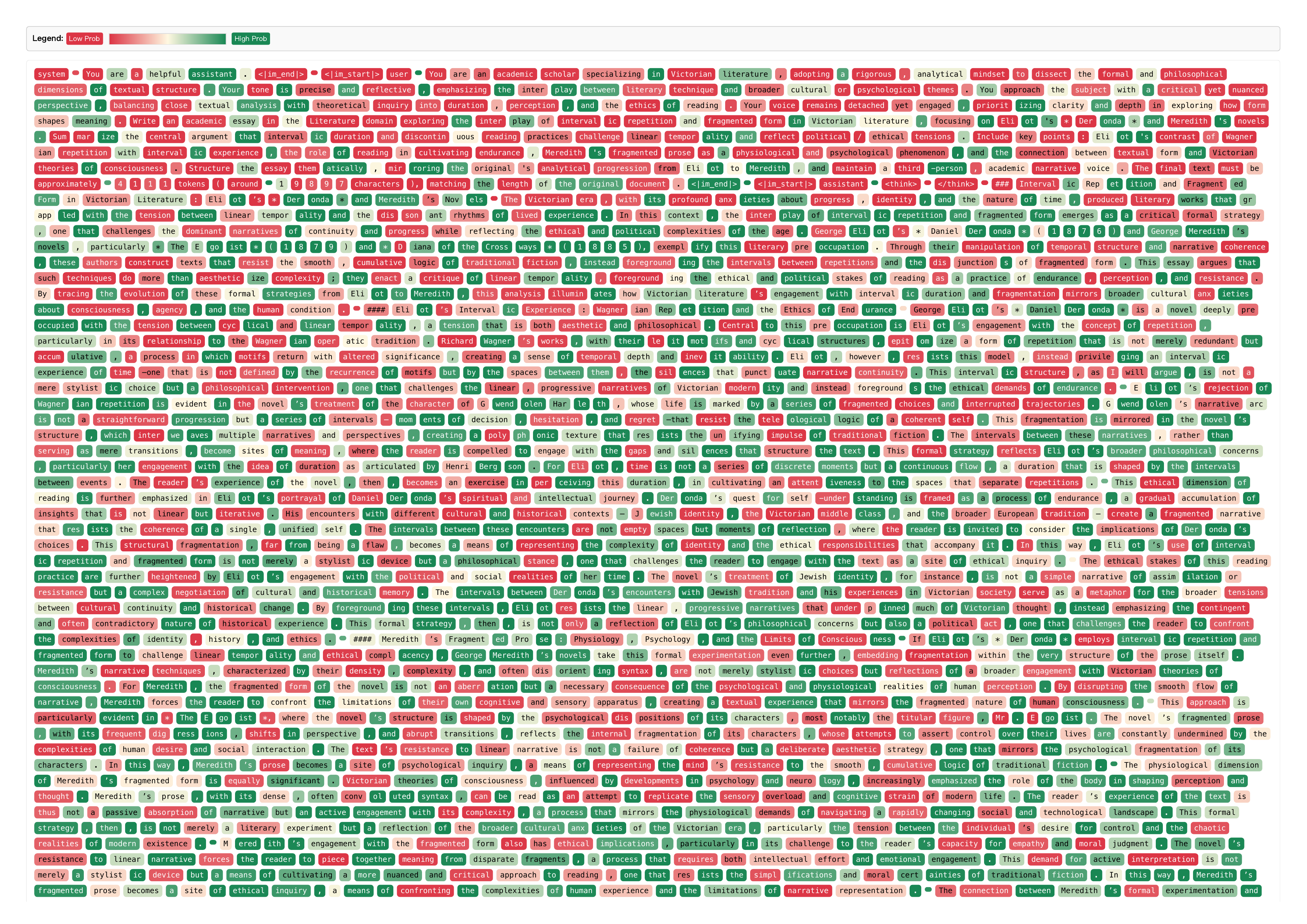}
        \vspace{-30pt} 
    \end{subfigure}    
    \begin{subfigure}{\linewidth}
        \centering
        \includegraphics[page=2,width=1.\linewidth]{figures/reverse_logprob_heatmap.pdf}
        \vspace{-85pt}
    \end{subfigure}
    \caption{Inference token likelihood heatmaps for models trained on \textbf{reverse answers}, where darker colors indicate tokens the model fails to predict confidently. The model trained on seed answers clearly finds its answer pattern easy to learn.}
    \label{fig:reverse_logprob_heatmap}
\end{figure*}

\begin{table*}[t]
\centering
\renewcommand{\arraystretch}{1.0}
\setlength{\tabcolsep}{2pt}
\resizebox{1.\linewidth}{!}{
\begin{tabular}{lccccc}
\toprule
\textbf{Quality Band} & \textbf{Rubric Score Range} & \textbf{Tier-specific Min.\ Thresholds} & \textbf{\#Docs (M)} & \textbf{\#Tokens (B)} & \textbf{Avg.\ Tok.\ Len (K)} \\
\midrule
Excellent & 55--60 & Read.$\ge$5, Applic.$\ge$5, H-Touch$\ge$4 & 0.29 & 1.10 & 3.75 \\
Seed      & 45--54 & Read.$\ge$5, Applic.$\ge$4, H-Touch$\ge$3 & 1.07 & 3.15 & 2.94 \\
Usable    & 31--44 & Read.$\ge$3, Applic.$\ge$3, H-Touch$\ge$2 & 12.14 & 12.53 & 1.03 \\
Unusable  & $\le$30 or any Readability$\le$2 & — & 1.50 & 1.62 & 1.08 \\
\bottomrule
\end{tabular}}
\vspace{-3pt}
\caption{The corpus quality bands and their statistics: Rubric Score Range denotes the total rubric score interval for each band; Tier-specific Min. Requirements list the minimum thresholds for the rubric scores across three weighted tiers (Readability, Applicability and Human-touch); \#Docs is the number of documents (in thousands); \#Tokens is the total token count (in billions); Avg. Tok. Len is the average document length (in thousands of tokens).}
\label{tab:quality_bands}
\vspace{-3pt}
\end{table*}

\section{Detail Process of Seed Document Construction}\label{sec:appendix_details_seed_construction}
\begin{itemize}[leftmargin=0pt,label={},itemsep=-2pt,topsep=2pt]
\item \textbf{Web Corpus}
We start from the 627B tokens SlimPajama corpus \cite{cerebras2023slimpajama}—a cleaned, deduplicated RedPajama derivative \cite{weber2024redpajama}, comprising 590M documents from seven sources: StackExchange, GitHub, CommonCrawl, C4, Books, arXiv, and Wikipedia.

\item \textbf{Tokenization and Chunking}
The web corpus is then tokenized with the Qwen3 tokenizer and, following \citet{abdin2024phi,xu2024magpie}, split into segments of 200–4k tokens to preserve the common length distribution.

\item \textbf{Source Sampling}
Following \citet{chen-etal-2024-dog}, we discard sources that lack HSS content (StackExchange, GitHub). Among the remaining five sources—CommonCrawl, C4, Books, arXiv, and Wikipedia—we retain all Books, arXiv, and Wikipedia documents due to their typical high quality, and randomly sample 10\% from the crawler-based C4 and CommonCrawl to balance source composition. This step yields an initial corpus of 30M documents.

\item \textbf{Heuristic Filtering}
Next, we process the initial corpus with the \texttt{Fineweb} toolkit%
\footnote{\url{https://github.com/huggingface/datatrove/blob/main/examples/fineweb.py}}, %
which sequentially applies the following heuristic filters:
\begin{enumerate}[leftmargin=13pt,itemsep=0pt,topsep=2pt]
    \item a fastText language classifier to filter out non-English text, using  the lid.176\footnote{\url{https://dl.fbaipublicfiles.com/fasttext/supervised-models/lid.176.bin}} backend model with a confidence threshold of~0.9 \cite{joulin2016fasttext};
    \item {GopherRepetitionFilter}, {GopherQualityFilter} \citep{rae2021scaling}, {C4QualityFilter} \citep{raffel2020exploring}, and {FineWebQualityFilter} \citep{penedo2024fineweb} to remove heuristically low-quality text.
    \item a MinHash deduplicator \citep{broder1997resemblance} to de-duplicate documents, configured with {n\_grams}=5, {num\_buckets}=14, and {hashes\_per\_bucket}=8 to balance recall and false positives.
\end{enumerate}

\item \textbf{Domain Classification}
Following the official QS subject taxonomy\footnote{\url{https://www.topuniversities.com/subject-rankings}}, we partition HSS into 14 core domains: philosophy, economics, law, politics, sociology, healthcare, geography, education, sports, literature, history, management, arts, and psychology. We then use the Qwen3-30B-A3B model to identify documents that fall into these HSS domains, yielding 2.09 million documents.

\item \textbf{Quality Rating}
We introduce 12 expert-crafted quality rubrics for LLM rating of HSS texts. Each rubric is rate 1–5 (definitions in Figure~\ref{fig:hss_quality_filter_part1_prompt},\ref{fig:hss_quality_filter_part2_prompt},\ref{fig:hss_quality_filter_part3_prompt}) and grouped into three weighted tiers:
\begin{itemize}[leftmargin=10pt, itemsep=0pt,topsep=2pt]
\item Readability: grammar, coherence, content accuracy, domain relevance (weight 0.5 each, maximum total 10).
\item Applicability: tone\& expression, knowledge depth, vocabulary richness, genre focus (weight 1.0 each; maximum total 20).
\item Human-touch: thematic Depth, emotionality, literary diversity, humanities creativity (weight 1.5 each; maximum total 30).
\end{itemize}
The overall rubric score ranges from 0 to 60. Based on tier-specific minimum requirements, we partition the corpus into four quality bands (see Table~\ref{tab:quality_bands}). We set the seed-document threshold at a rubric score of 45, as this balances corpus quality—perfect readability ($\ge$5), good applicability ($\ge$4), and moderate human-touch ($\ge$3)—with scale. A higher threshold would substantially reduce data volume and domain coverage, which is detrimental to subsequent instruction backtranslation. After this stage, 1.07 million documents remain.

\item \textbf{Text Refinement}
Even after multi-step processing, residual noise, redundancy, or weak expression may persist and cannot be remedied by filtering alone. We therefore employ LLM-based text refinement to achieve: i) content cleaning (removing crawl artifacts and irrelevant redundancy); ii) content fidelity (preserving core information, structure, and “human-touch”); and iii) expression optimization (improving coherence, fluency, and accuracy).

\item \textbf{Refinement Judge}
We use an LLM to judge whether a refined document meets the three goals above, and accordingly decide whether it qualifies as a curated seed document.
\end{itemize}

\begin{table*}[t]
\centering
\renewcommand{\arraystretch}{0.95}
\begin{tabular}{lcccc}
\toprule
\textbf{Setting} &
\textbf{Q\&A Alignment} &
\multicolumn{2}{c}{\textbf{Score Range}}\\
\cmidrule(lr){3-4}
& & 45--54 & 55--60 \\
\midrule
Filtered-Only      & 938 (93.8\%)  & 7917 (79.2\%)  & 2083 (20.8\%)  \\
Filtered+Refined   & 959 (95.9\%)  & 7758 (77.6\%)  & 2242 (22.4\%)  \\
$\Delta$ Gain      & +21 (2.1\%) & -159 (-1.6\%) & +159 (1.6\%) \\
\bottomrule
\end{tabular}
\vspace{-3pt}
\caption{
Comparison of Q\&A alignment and quality rubric score distributions before (Filtered-Only) and after text refinement (Filtered+Refined). Q\&A Alignment denotes the documents that pass GPT-4.1’s Q\&A consistency check; Score ranges 45–54 and 55–60 indicate the documents whose quality rubric scores fall within these bands.
}
\label{tab:refine_quality}
\vspace{-15pt}
\end{table*}

\section{Necessity of Text Refinement}~\label{sec:necessity_refinement}
Since data filtering alone is insufficient to produce clean seed documents, we introduce LLM-based text refinement. To validate its necessity, we randomly sample 1k instances to compare Q\&A alignment before (Filtered-Only) and after text refinement (Filtered+Refined), and further evaluate quality rubric scores on 10k instances.

\begin{itemize}[leftmargin=11pt, itemsep=0pt,topsep=2pt]
\item \textbf{Better Q\&A alignment}: We backtranslate instructions from seed documents in the Filtered-only and Filtered+Refined sets, and use GPT-4.1 to assess Q\&A alignment. The Filtered+Refined set exhibits a markedly lower rejection rate, reducing 21 Q\&A misaligned instances in the 1k samples, indicating that text refinement improves the fidelity of instruction backtranslation.
\item \textbf{Better Quality distribution}: Table \ref{tab:refine_quality} reports the distribution of quality rubric scores between Filtered-only  and Filtered+Refined. After text refinement, the proportion of excellent documents (score range of 55–60) increases substantially, adding 159 such documents in the 10k samples.
\end{itemize}

\section{Domain Distribution of HSS-Synth Dataset}
\vspace{-2pt}
To demonstrate the domain diversity of the instruction-tuning dataset synthesized by HSS-Synth, we present statistics on its domain distribution in Table~\ref{tab:domain_distribution}, with a total of 237,340 samples.

\begin{table}[t]
\centering
\begin{tabular}{l r r}
\toprule
Domain & \#Docs & Proportion (\%) \\
\midrule
History     & 56,241 & 23.70 \\
Arts        & 35,390 & 14.91 \\
Politics    & 28,966 & 12.20 \\
Literature  & 27,664 & 11.66 \\
Philosophy  & 15,204 &  6.41 \\
Sociology   & 13,191 &  5.56 \\
Economics   & 13,123 &  5.53 \\
Law         &  9,217 &  3.88 \\
Sports      &  8,548 &  3.60 \\
Psychology  &  8,183 &  3.45 \\
Management  &  7,662 &  3.23 \\
Education   &  7,641 &  3.22 \\
Healthcare  &  5,129 &  2.16 \\
Geography   &  1,181 &  0.50 \\
\midrule
Total       & 237,340 & 100.00 \\
\bottomrule
\end{tabular}
\caption{Domain distribution of the HSS-Synth dataset.}
\label{tab:domain_distribution}
\end{table}

\section{Model Training Details}~\label{sec:appendix_training}
Our model training is conducted on the \texttt{Llama-factory}\footnote{https://github.com/hiyouga/LLaMA-Factory}~\cite{zheng-etal-2024-llamafactory} using 64x NVIDIA H800 GPUs. We trained the Qwen3-8B-Base model for three epochs, each taking approximately 1 hour, totaling about 3 hours. The checkpoint with the lowest validation loss was selected as the final model. Detailed training hyperparameters are listed in Table~\ref{tab:hyperparams}.

\begin{table}[h]
  \centering
  \vspace{-7pt}
  \renewcommand{\arraystretch}{0.9}
  \begin{tabular}{l c}
    \toprule
    \textbf{Hyperparameter} & \textbf{Value} \\
    \midrule
    Batch size        & 512   \\
    Learning rate     & 7e$^{-6}$ \\
    Maximum length    & 5120  \\
    Epochs            & 3     \\
    Scheduler         & Cosine \\
    Warmup ratio      & 0.03  \\
    Weight decay      & 0.1   \\
    Adam $\beta_{2}$  & 0.95  \\
    Precision         & BF16  \\
    Random seed       & 42    \\
    \bottomrule
  \end{tabular}
  \vspace{-2pt}
  \caption{Training hyperparameters used for Qwen3{-}8B{-}Base, Qwen2.5{-}14B, and Llama3.1{-}8B.}
  \label{tab:hyperparams}
  \vspace{-16pt}
\end{table}

\section{Case Study}
We provide a case from the HSS‑Synth synthesis pipeline in Table~\ref{tab:case_study}, comprising the seed document, reverse instruction, reverse answer, and teacher‑forced answer. This case shows that, unlike closed tasks with verified answers, HSS questions (i.e., reverse instruction) admit open‑ended responses, such as both the reverse and teacher‑forced answers. The reverse instruction presents a richly specified persona and faithfully reflects the content of the seed document. Compared with the reverse answer (433 tokens), the teacher‑forced answer (644 tokens) satisfies each requirement of the instruction more precisely—covering the essay core information (four skills of the Heart IQ model: emotional connection, embodiment, love, and authenticity), the essay organization, and the length constraints (413 tokens)—demonstrating TeachForceA’s strengths in supplying requisite information to reduce hallucination and leveraging structural and lexical cues to improve completeness and readability. Moreover, it preserves the tone and style of the seed document, yielding a more human‑touch quality.

\begin{table*}[t]
\tiny
\setlength{\tabcolsep}{2pt}
\begin{tabularx}{\textwidth}{lX}
\toprule
\textbf{Text Type} & \textbf{Text}  \\
\midrule
Seed Document & \detokenize{The Heart IQ Model consists of four skills that are developed together. A good analogy is a four legged stool. If each skill represents a leg on the stool, you can appreciate the need for all four legs to be of equal length or the stool becomes unstable and thus unusable. In Heart IQ, we've found a unique way to awaken these skills in our students through an organic, highly experiential approach.\n\nWhen we speak about opening the Heart, we are not referring to the pump that pushes blood around your body! Instead, we are talking about FEELING MORE. Feeling more connected to source, feeling more embodied, feeling more love from those you care about and feeling more power and authenticity in your expression to others.\n\nWhen your heart is closed, you will find it difficult to access your unique guidance system. Instead of hearing the subtle messages of your inner wisdom, you will only hear the loud noise of your conditioned mind. When you learn to listen to your heart (through stillness and a process we call 'Tracking'), you will develop the ability to tell the difference between your authentic longings and those that were given to you by your family, education and culture. Through practice, your true purpose and life's direction is revealed.\n\nTo know your deepest heart's longing is only part of the journey. What's required next is to follow your heart and take inspired action. In Heart IQ\u2122 you will develop the willingness, courage and trust required to take fearless action so that you can manifest the life of your dreams.\n\nOne of the most rewarding aspects of learning Heart IQ\u2122 is how it transforms your relationships with others. Learning to communicate from the heart is a vital life skill that serves all relationships, whether it's with your partner, your customers or your children. Most of us were never taught the critical steps to healthy heart centred communication and conflict resolution. With practice you will develop the skill to express yourself in a way that really lands in the hearts and minds of others while having the tools to resolve differences and create harmony with those you love.}\\
\midrule
Reverse Instruction & \detokenize{You are a compassionate guide specializing in emotional intelligence, dedicated to helping readers unlock their inner wisdom. Your stance prioritizes holistic self-discovery and authentic living. You approach the topic with empathetic curiosity, balancing practical advice with spiritual insight. Your tone is reassuring and empowering, fostering a sense of possibility while maintaining intellectual rigor.\nWrite a Psychology self-help essay explaining the Heart IQ Model's four interdependent skills using the four-legged stool analogy. Emphasize the concept of 'opening the heart' as emotional connection, embodiment, love, and authenticity. Include the process of 'Tracking' to distinguish authentic desires, the importance of inspired action for manifesting goals, and the role of heart-centered communication in relationships. Organize the essay with a problem-solution structure, starting with the consequences of a closed heart, followed by the model's methodology, and concluding with relational benefits. The final text must be approximately 413 tokens (around 2130 characters), matching the length of the original document.} \\
\midrule
Reverse Answer & \detokenize{**The Heart IQ Model: A Four-Legged Stool for Emotional Resilience**  \n\nA closed heart\u2014marked by emotional detachment, fear, or self-censorship\u2014leaves us isolated, reactive, and disconnected from our true selves. Without emotional openness, we struggle to form meaningful relationships, pursue authentic goals, or navigate life\u2019s challenges with grace. This imbalance weakens our capacity for empathy, creativity, and joy, trapping us in cycles of dissatisfaction. The *Heart IQ Model* offers a transformative framework to rebuild emotional stability by cultivating four interdependent skills, much like a four-legged stool: *opening the heart*, *tracking desires*, *inspired action*, and *heart-centered communication*.  \n\n**1. Opening the Heart: The Foundation of Emotional Connection**  \nThe first leg of the stool is *opening the heart*, which involves embracing emotional vulnerability, embodiment (physical and emotional presence), love, and authenticity. A closed heart numbs us to our needs and others\u2019 experiences, but opening it allows us to feel deeply, connect authentically, and align with our values. This process begins with self-compassion\u2014acknowledging pain without judgment and allowing space for healing. When we open our hearts, we become more attuned to our inner wisdom, creating a foundation for the other skills.  \n\n**2. Tracking: Distinguishing Authentic Desires**  \nThe second leg, *tracking*, is the practice of observing thoughts, emotions, and bodily sensations to discern what truly resonates with our authentic self. In a world of external noise, this skill helps us differentiate between societal expectations and inner calling. For example, a person might track their energy levels after a social interaction to identify whether it aligns with their need for connection or drains their spirit. Tracking is not about suppression but about cultivating awareness, enabling us to make choices that honor our deepest desires.  \n\n**3. Inspired Action: Manifesting Goals with Purpose**  \nThe third leg, *inspired action*, bridges intention and manifestation. When we act from a place of heart-centered clarity, our efforts feel aligned with our values, fostering resilience and joy. This is not about forced productivity but about responding to life\u2019s callings with curiosity and courage. For instance, someone passionate about art might begin a creative project not out of obligation but because it feels like a natural extension of their being. Inspired action transforms abstract dreams into lived experiences, reinforcing our connection to purpose.  \n\n**4. Heart-Centered Communication: Building Authentic Relationships**  \nThe final leg, *heart-centered communication*, emphasizes presence, honesty, and empathy in interactions. It requires listening without judgment, speaking from a place of vulnerability, and honoring the emotional needs of others. This skill strengthens relationships by fostering trust and mutual growth. When we communicate from the heart, we create space for others to do the same, building communities rooted in compassion and understanding.  \n\n**The Interdependence of the Four Legs**  \nLike a stool, the Heart IQ Model thrives when all four legs are strong. A closed heart hinders tracking, as fear clouds self-awareness. Without tracking, inspired action lacks direction. And without heart-centered communication, even the most authentic goals remain unshared. Together, these skills form a dynamic system that supports emotional resilience, authenticity, and connection.  \n\n**Relational Benefits: A Heart-Open World**  \nWhen we nurture these skills, relationships deepen. We become more patient, empathetic, and present, fostering environments where others feel safe to open their hearts too. This creates a ripple effect\u2014individual healing expands into collective well-being.  \n\nIn closing, the Heart IQ Model invites us to rebuild our emotional lives not through force, but through gentle, consistent practice. By opening our hearts, tracking our truths, acting with purpose, and communicating with authenticity, we cultivate a life of meaning, connection, and inner peace. The stool stands firm when all legs are nurtured, reminding us that emotional intelligence is not a destination but a journey of continual growth.} \\
\midrule
Teacher-forced Answer & \detokenize{When the heart is closed, the soul\u2019s compass falters. A rigid, guarded heart muffles the quiet whispers of intuition, leaving individuals adrift in a sea of external validation and conditioned fears. Without emotional connection, embodiment, love, and authenticity, life becomes a series of fragmented attempts to fill an unspoken void. Relationships fray, goals feel hollow, and the self dissolves into a shadow of what it might be. This is the cost of a heart unopened\u2014a dissonance between inner truth and outer expression, where authenticity is buried under the weight of societal expectations.  \n\nThe Heart IQ Model offers a pathway to reclamation, anchored in the metaphor of a four-legged stool. Each leg\u2014emotional connection, embodiment, love, and authenticity\u2014must be equally strong to sustain balance. Emotional connection is the foundation, grounding us in the present through mindful presence. Embodiment, the second leg, bridges the mind and body, allowing us to feel rather than merely think. Love, the third, expands our capacity to give and receive without condition, while authenticity, the fourth, ensures our actions align with our deepest values. Together, they form a stable platform for holistic growth.  \n\nTo unlock this balance, the practice of *Tracking* becomes essential. This is not mere introspection but a deliberate act of stillness, where one distinguishes between the clamor of external influences and the quiet resonance of authentic desire. By sitting with discomfort and curiosity, we learn to recognize the difference between cultural scripts and the soul\u2019s true yearning. This clarity paves the way for *inspired action*\u2014not driven by fear or obligation, but by a vibrant alignment with purpose. When we act from this space, goals transform from distant ideals into lived realities, fueled by courage and trust.  \n\nYet the true alchemy of Heart IQ lies in its relational impact. Heart-centered communication, rooted in empathy and vulnerability, dissolves barriers and fosters profound connection. It is not about perfection but presence: listening without judgment, speaking with integrity, and resolving conflict as a collaborative dance rather than a battle. Such practices mend fractured relationships, creating spaces where love can flourish unencumbered by fear.  \n\nTo open the heart is to embrace life\u2019s full spectrum\u2014its joys, its sorrows, its infinite possibilities. It is a journey of becoming, not a destination, where each step deepens our capacity to live authentically and connect meaningfully. In this light, the Heart IQ Model is not merely a tool but a testament to the human spirit\u2019s resilience\u2014a reminder that when we dare to open, we unlock the wisdom that has always dwelled within.} \\
\bottomrule
\end{tabularx}
\caption{
A case from the HSS‑Synth synthesis pipeline, comprising the seed document, reverse instruction, reverse answer, and teacher‑forced answer. 
}\label{tab:case_study}
\end{table*}

\section{Prompt Template} \label{sec:prompt_template}
\begin{figure*}[!t]
\centering
\includegraphics[width=1.\linewidth]{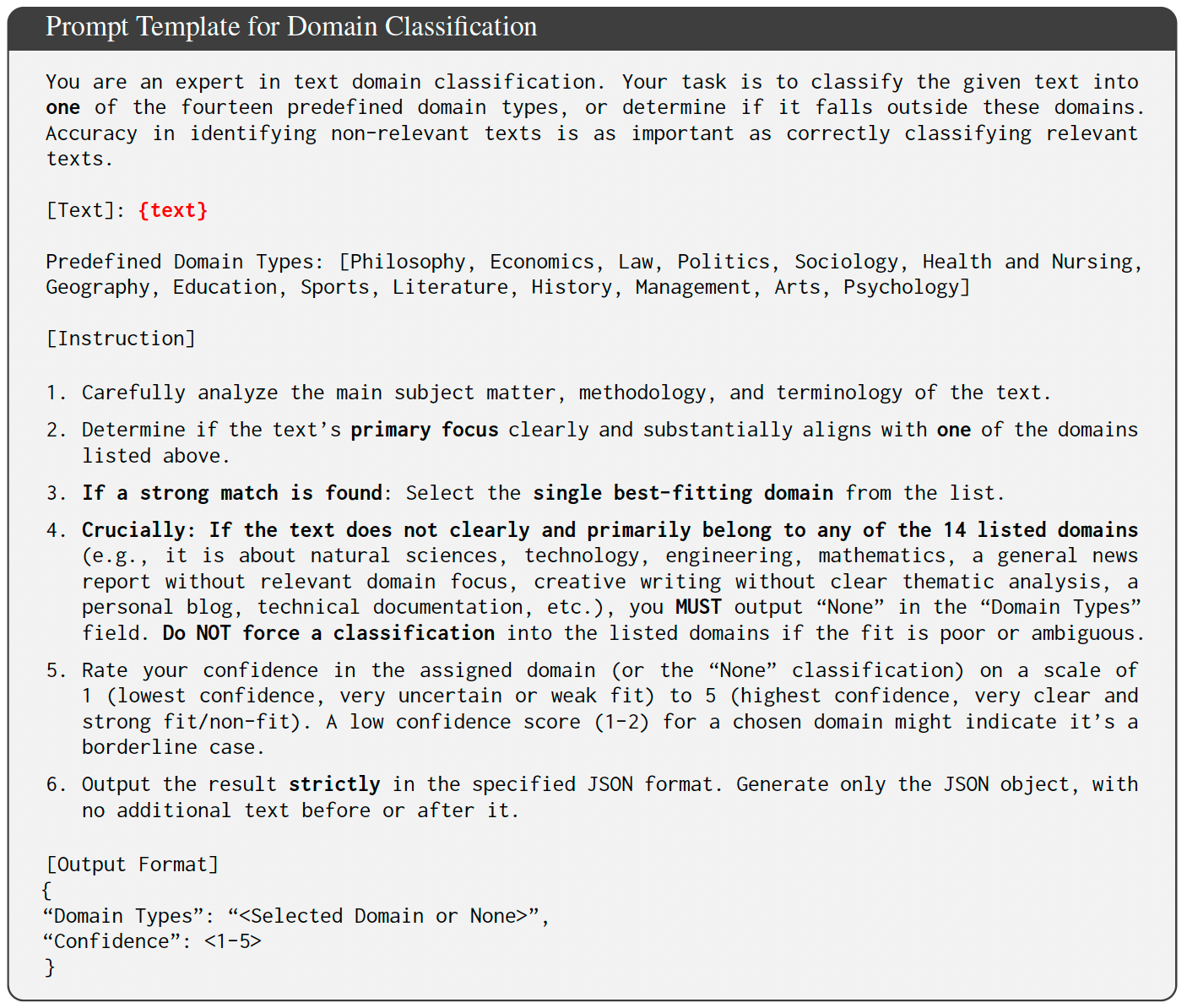}
\caption{
Prompt template for domain classification to classify documents into HSS domains.
}
\label{fig:hss_domain_cls_prompt}
\end{figure*}

\begin{figure*}[!t]
\centering
\includegraphics[width=1.\linewidth]{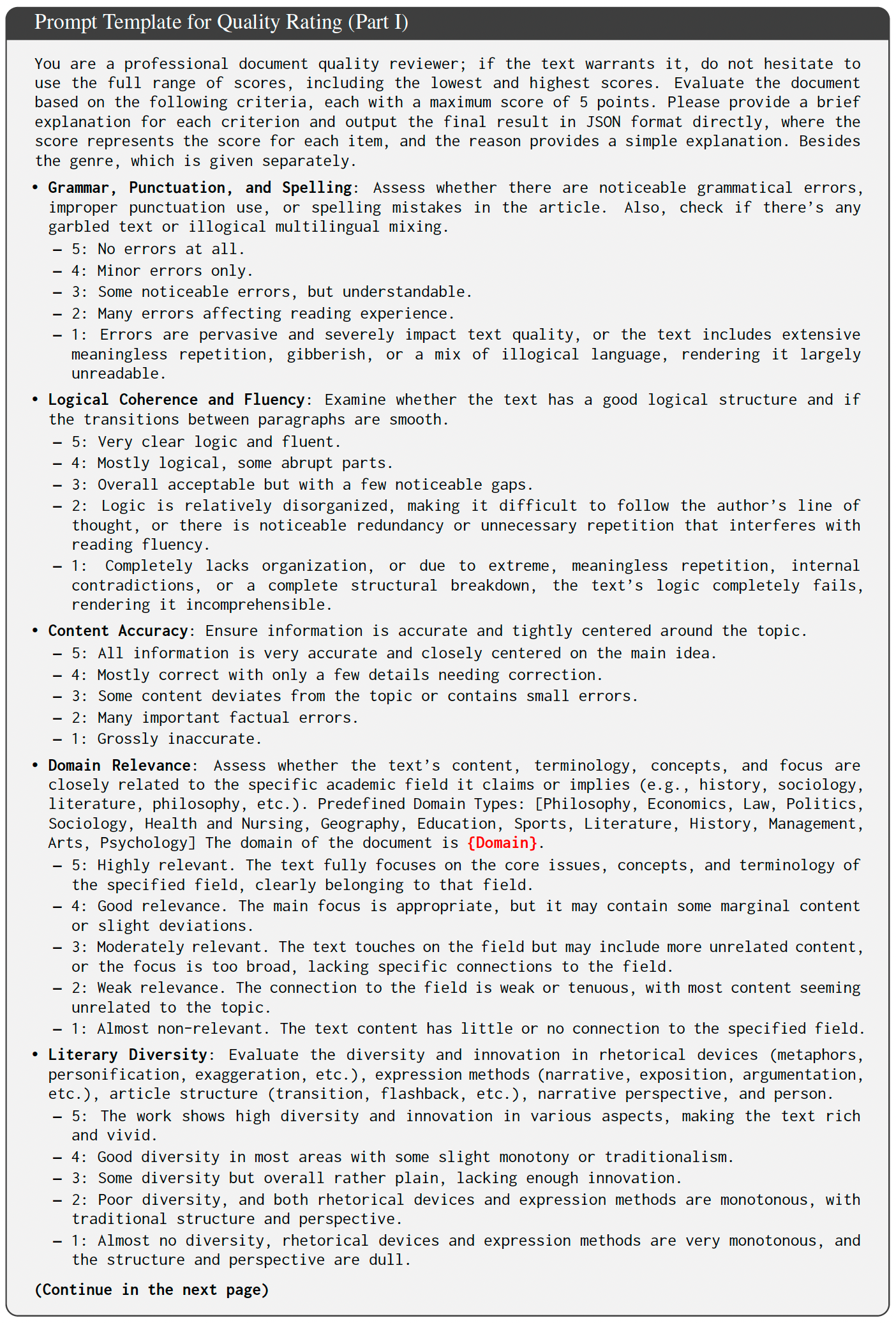}
\caption{
Prompt template for quality rating that assesses HSS documents using expert-crafted rubrics (Part I).
}
\label{fig:hss_quality_filter_part1_prompt}
\end{figure*}

\begin{figure*}[!t]
\centering
\includegraphics[width=1.\linewidth]{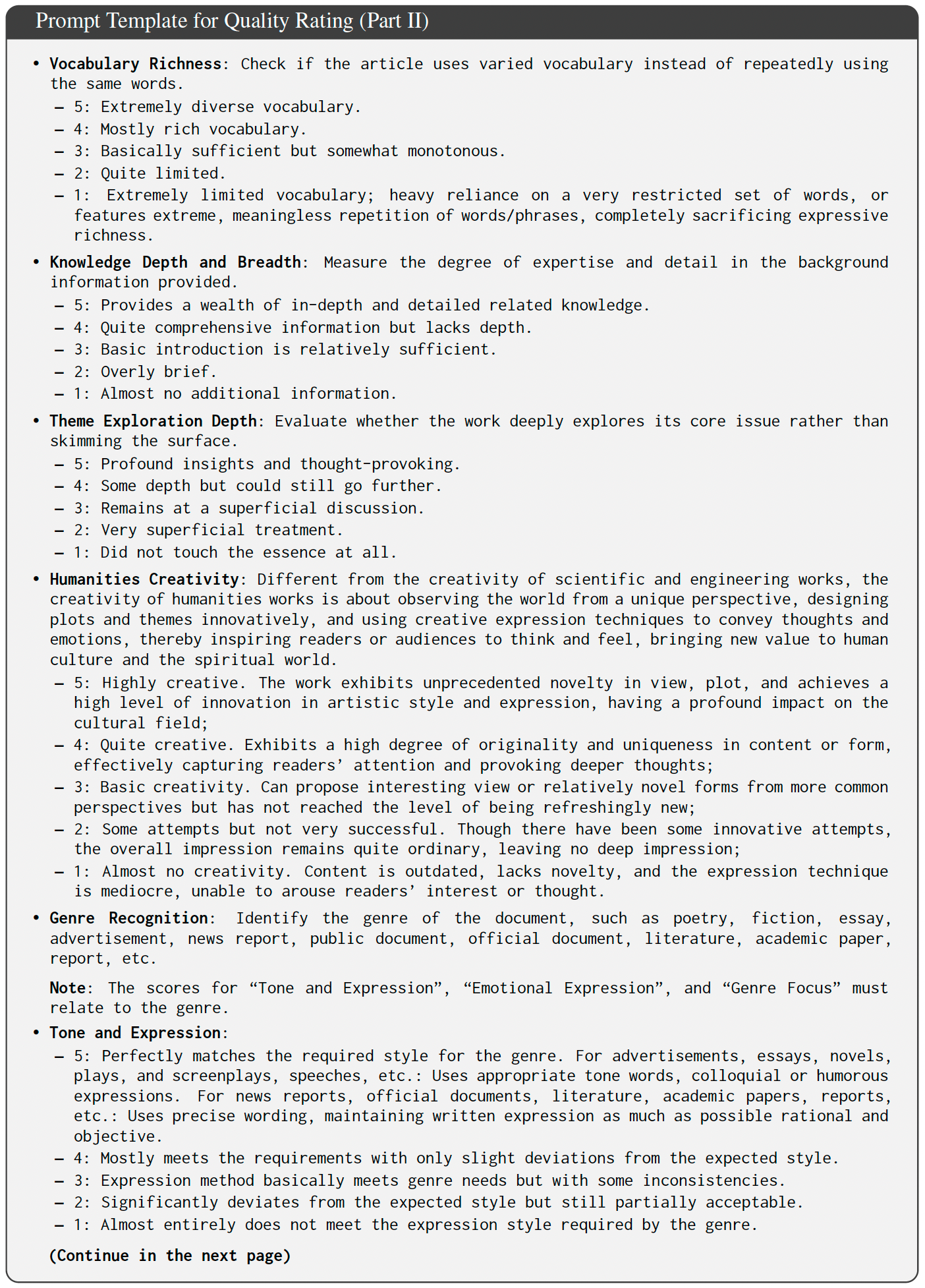}
\caption{
Prompt template for quality rating that assesses HSS documents using expert-crafted rubrics (Part II).
}
\label{fig:hss_quality_filter_part2_prompt}
\end{figure*}

\begin{figure*}[!t]
\centering
\includegraphics[width=1.\linewidth]{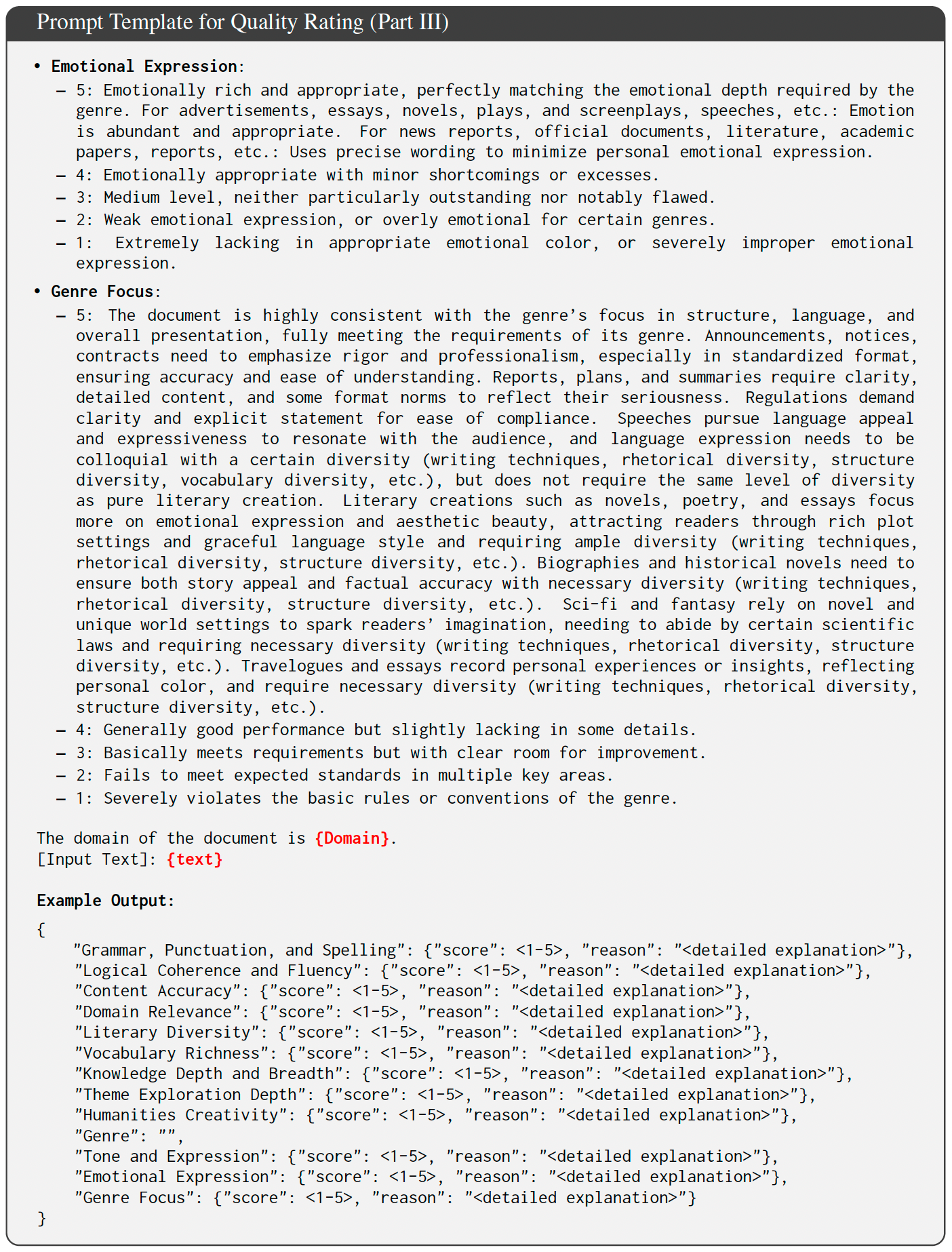}
\caption{
Prompt template for quality rating that assesses HSS documents using expert-crafted rubrics (Part III).
}
\label{fig:hss_quality_filter_part3_prompt}
\end{figure*}

\begin{figure*}[!t]
\centering
\includegraphics[width=1.\linewidth]{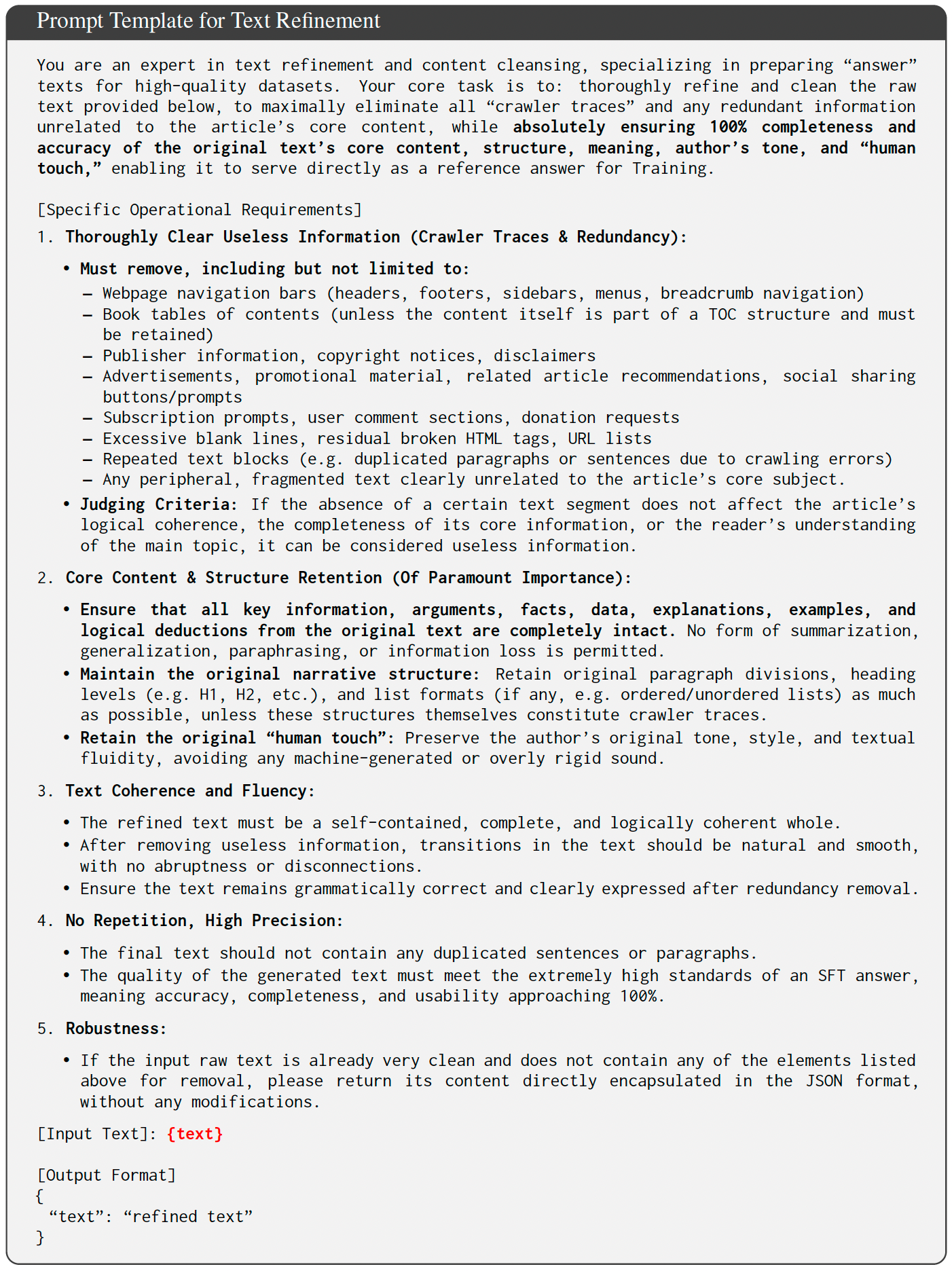}
\caption{
Prompt template for text refinement to polish seed documents.
}
\label{fig:hss_text_refinement_prompt}
\end{figure*}

\begin{figure*}[!t]
\centering
\includegraphics[width=1.\linewidth]{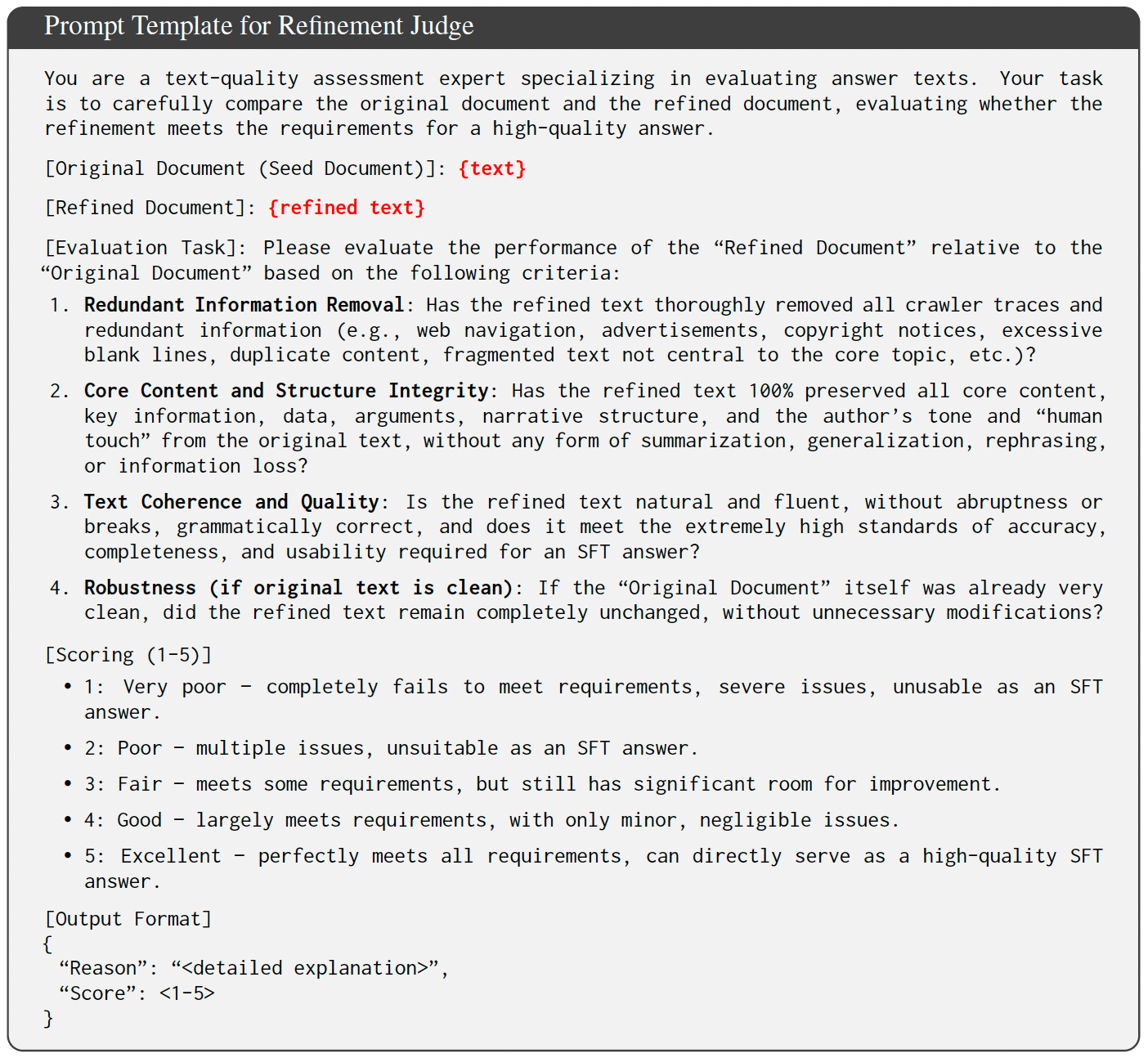}
\caption{
Prompt template for a refinement judge that determines whether text refinement is successful.
}
\label{fig:hss_refinement_judgement_prompt}
\end{figure*}

\begin{figure*}[!t]
\centering
\includegraphics[width=1.\linewidth]{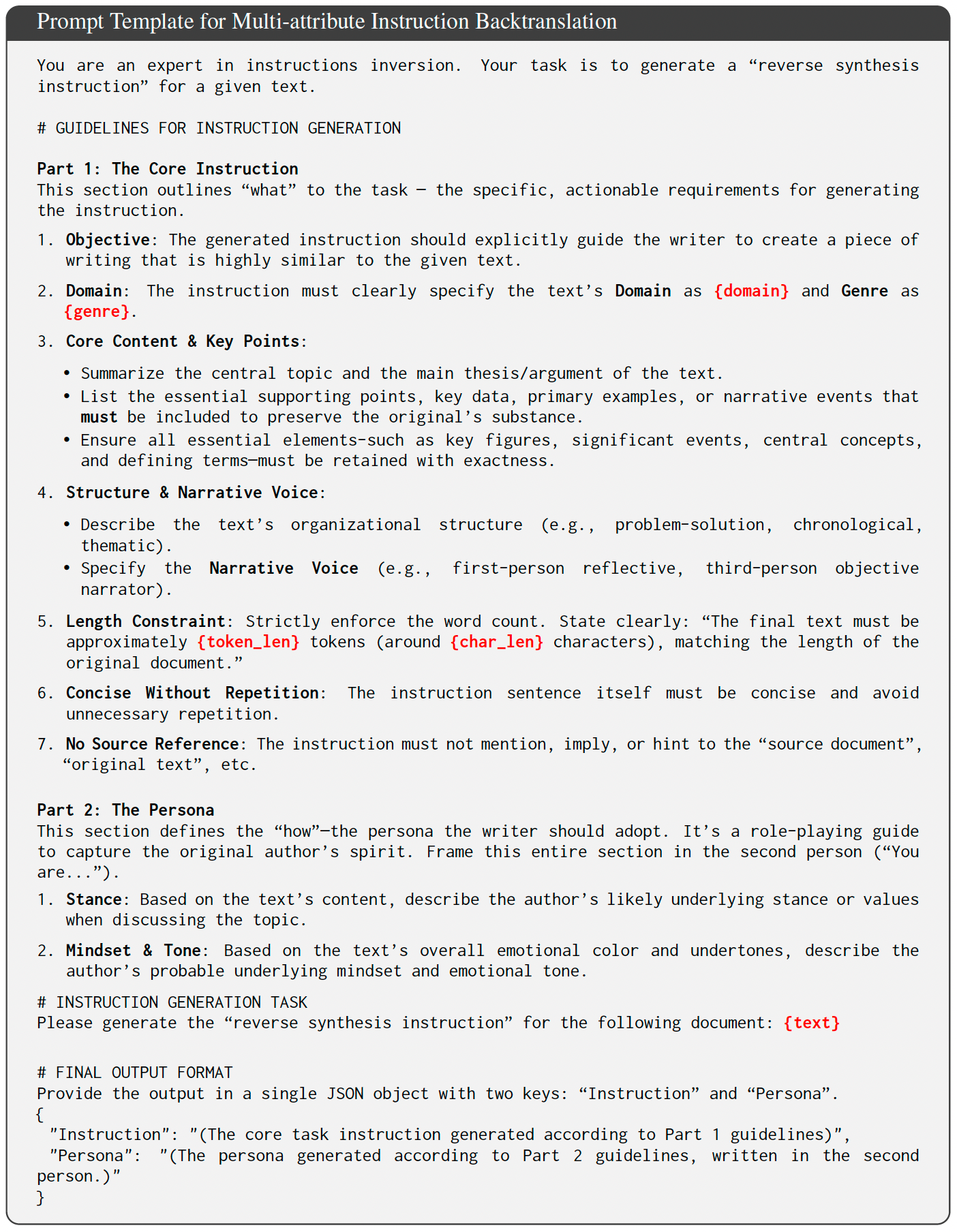}
\caption{
Prompt template for multi-attribute instruction backtranslation, which backtranslates instructions from seed documents while attaching multiple attributes.
}
\label{fig:hss_instruction_inversion_prompt}
\end{figure*}

\begin{figure*}[!t]
\centering
\includegraphics[width=1.\linewidth]{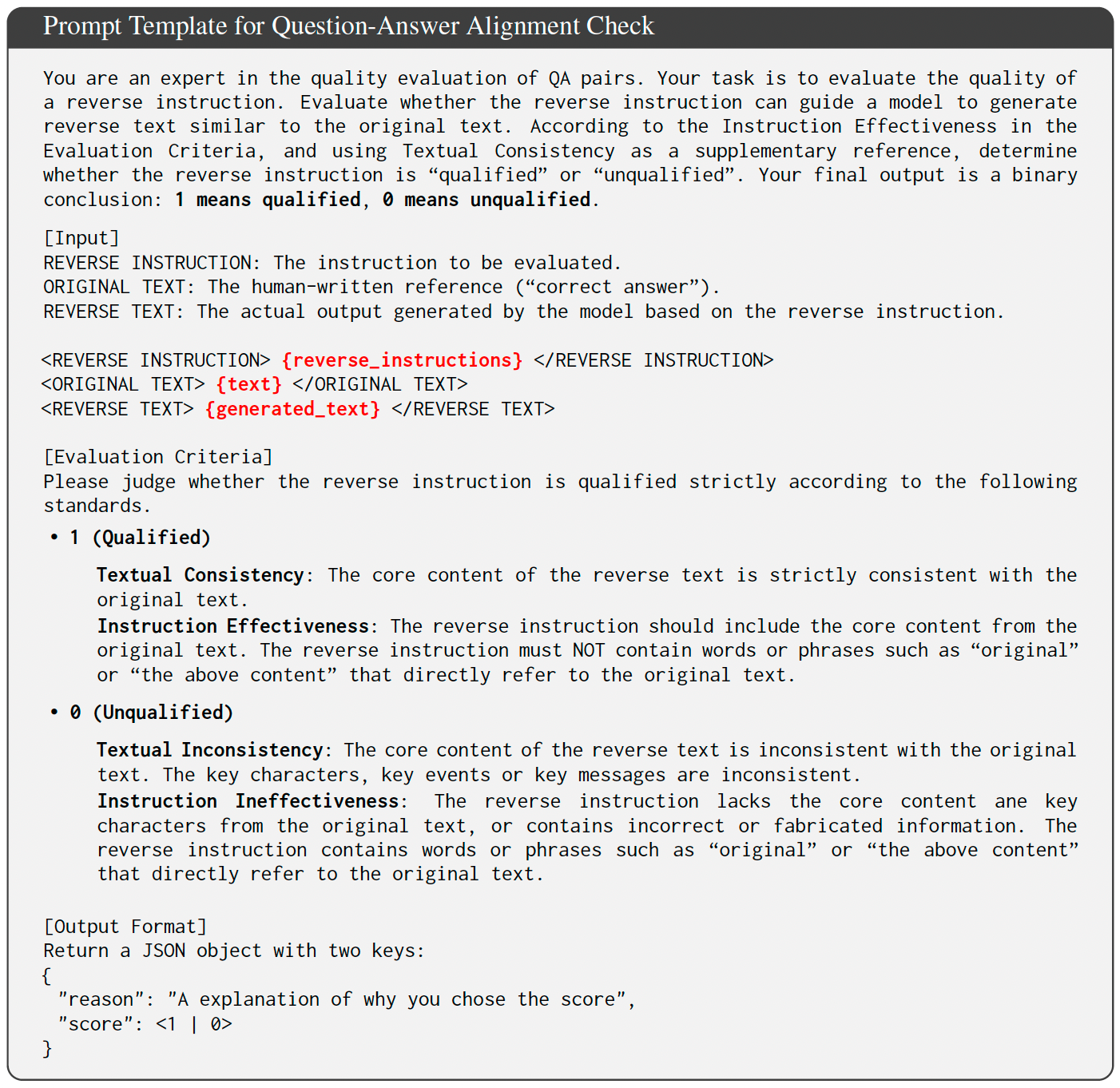}
\caption{
Prompt template for question-answer alignment check used to verify that the reverse instructions faithfully reproduce the seed document.
}
\label{fig:hss_qa_consistency_check_prompt}
\end{figure*}

\begin{figure*}[!t]
\centering
\includegraphics[width=1.\linewidth]{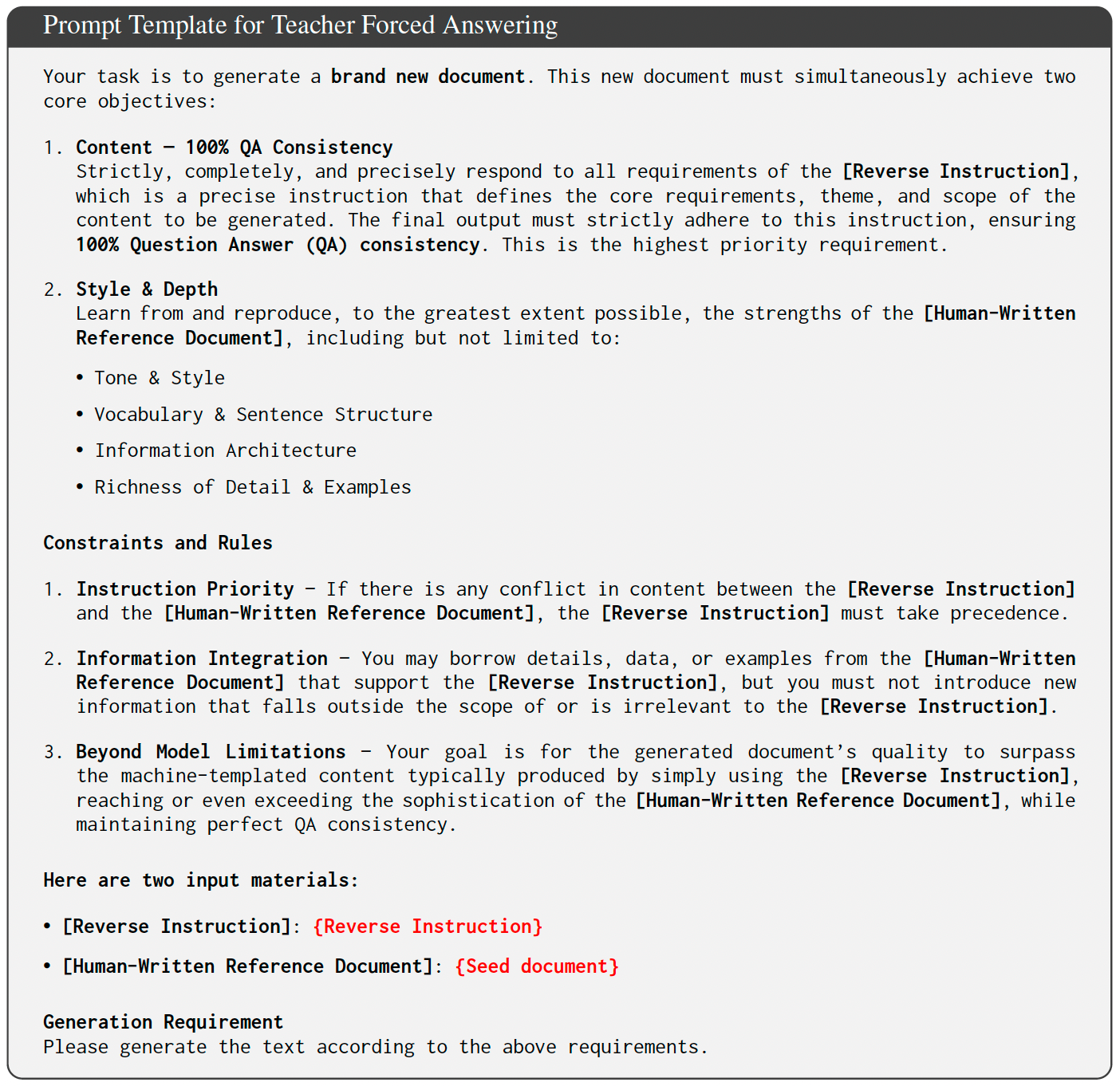}
\caption{
Prompt template for teacher-forced answering, which produces answers anchored to seed documents that exceed the synthesis model’s response limits.
}
\label{fig:hss_teacher_forced_answering_prompt}
\end{figure*}

This section summarizes the prompt templates used at each stage.
Figure \ref{fig:hss_domain_cls_prompt} shows the prompt template for domain classification to classify seed documents into HSS domains.
Figures \ref{fig:hss_quality_filter_part1_prompt}, \ref{fig:hss_quality_filter_part2_prompt}, and \ref{fig:hss_quality_filter_part3_prompt} show the prompt template for quality rating that assesses HSS documents using expert-crafted rubrics.
Figure \ref{fig:hss_text_refinement_prompt} shows the prompt template for text refinement to polish seed documents.
Figures \ref{fig:hss_text_refinement_prompt} and \ref{fig:hss_refinement_judgement_prompt} show the prompt template for refinement judgement that determines whether text refinement is successful.
Figure \ref{fig:hss_instruction_inversion_prompt} shows the prompt template for multi-attribute instruction backtranslation, which backtranslates instructions from seed documents while attaching multiple attributes.
Figure \ref{fig:hss_qa_consistency_check_prompt} shows the prompt template for the question-answer alignment check used to verify that the reverse instructions faithfully reproduce the seed document.
Figure \ref{fig:hss_teacher_forced_answering_prompt} shows the prompt template for teacher-forced answering, which produces answers anchored to seed documents that exceed the synthesis model’s response limits.

\end{document}